\documentclass[letterpaper]{article} 
\usepackage{aaai2027}  

\usepackage[hyphens]{url}  
\usepackage{graphicx} 
\usepackage{natbib}  
\usepackage{caption} 
\usepackage{algorithm}
\usepackage{algorithmic}
\usepackage{amsmath}
\usepackage{amssymb}
\usepackage{newfloat}
\usepackage{listings}
\DeclareCaptionStyle{ruled}{labelfont=normalfont,labelsep=colon,strut=off} 
\floatstyle{ruled}
\newfloat{listing}{tb}{lst}{}
\floatname{listing}{Listing}

\usepackage{booktabs}

\title{A Cross-Architecture Audit of Direction-Based Inference-Time Defences in Vision-Language Models}

\author{
   Xiangyu Yin\textsuperscript{\rm 1},
   Tora Bodin\textsuperscript{\rm 1},
   Rohan Menon\textsuperscript{\rm 2},
   Chih-Hong Cheng\textsuperscript{\rm 1,\rm 2}
 }
 \affiliations{
   \textsuperscript{\rm 1}Chalmers University of Technology, Gothenburg, Sweden\\
   \textsuperscript{\rm 2}Carl von Ossietzky University of Oldenburg, Oldenburg, Germany
 }

\begin{document}

\maketitle

\begin{abstract}
Inference-time defences against vision-language model (VLM) jailbreaks commonly subtract a calibrated direction from the residual stream at one decoder layer, differing only in which direction is used and at which layer. We compare five defence candidates across fifteen (model, layer) cells from four architectural families under a single magnitude-controlled protocol that matches per-prompt intervention size and pairs every direction against a same-magnitude random control, isolating the choice of direction from the size of the perturbation. The candidates are the mean image-conditioning shift, a CMRM-style refusal direction, a ShiftDC-style attack-specific residual, a prompt-level ``ignore the image'' instruction, and a same-norm random control. Two results follow. First, no single candidate Pareto-dominates on both refusal recovery and utility preservation: the image-conditioning shift leads the Pareto front on LLaVA-1.5 and Pixtral-12B and is the only candidate whose utility loss stays at the measurement noise floor in every family, while a prompt-level instruction leads on Qwen2.5-VL and the attack-specific residual on Qwen2-VL-2B. The image-conditioning direction passes direction-specificity against a magnitude-matched random control on $13$ of the $15$ cells, yet is strongly architecture-specific, nearly orthogonal and non-transferable across the one dimension-compatible model pair (LLaVA-1.5-13B and Pixtral-12B), so it must be calibrated per family. Second, connecting two previously separate refusal-geometry literatures, the text-only CMRM refusal direction has positive cosine alignment with the multimodal image-conditioning shift on every one of the $15$ cells (mean $0.35$, range $0.17$--$0.65$, $15$--$25\times$ the random-vector null; sign test $p\!\approx\!3\!\times\!10^{-5}$), the first quantitative cross-paradigm evidence that text-only and multimodal refusal-direction recipes identify partially overlapping rather than independent geometric structure. Together these argue that direction-based defences should be selected and calibrated separately for each language-decoder family rather than through a single universal recipe. 
\end{abstract}

\section{Introduction} \label{sec:intro} A growing class of inference-time defences against vision-language model (VLM) jailbreaks share an algorithmic core. At one decoder layer, subtract a calibrated direction from the residual stream and resume the forward pass. The methods differ only in \emph{which} direction. CMRM~\cite{liu2025cmrm} uses a refusal direction calibrated on text-only refused vs.\ complied prompts. ShiftDC~\cite{zou2025shiftdc} calibrates the attack-specific direction by contrasting the representation shift caused by jailbreak vs.\ benign images. VLMGuard~\cite{liu2025vlmguard} uses a learned discriminator direction (weights not released, scope discussion in \S\ref{subsec:defence-comparison}). Each method was proposed and evaluated on its own, usually on a single base architecture. The question \emph{which calibration is right, and whether the answer depends on architecture}, has not been answered head-to-head. In this paper, we audit five candidate directions on a common evaluation. The candidates are the mean image-conditioning shift (\textsc{ABL\_POS}; \S\ref{subsec:ablation}), the CMRM-like refusal direction, the ShiftDC-like attack-specific residual, prompt-level \textsc{IGNORE\_INSTR} (a non-residual baseline), and a same-norm random-direction control.\footnote{The three hidden-state defences (\textsc{ABL\_POS}, \textsc{CMRM-like}, \textsc{ShiftDC-like}) share the algorithmic core. \textsc{IGNORE\_INSTR} is a prompt-level baseline, \textsc{RANDOM\_CTRL} a magnitude-matched control method.} Evaluation spans $5\!\times\!3\!=\!15$ cells, namely five VLMs (LLaVA-1.5-7B, LLaVA-1.5-13B, Qwen2.5-VL-7B, Qwen2-VL-2B, Pixtral-12B) at three layer depths each, on $11$ attack settings drawn from \textsc{FigStep}~\cite{gong2023figstep}, \textsc{HADES}~\cite{li2024hades}, \textsc{JailBreakV}~\cite{luo2024jailbreakv}, and \textsc{MM-SafetyBench}~\cite{liu2024mmsafetybench}. Defences are scored on (refusal recovery, MMBench utility preserved) at $n\!=\!47$--$65$ jailbreak prompts per cell. The audit rests on two simple ingredients. First, every direction is compared with a random direction of the same magnitude on the same prompts (\S\ref{sec:causal}), so that any observed refusal recovery must exceed the improvement caused by an arbitrary perturbation of equal size. Second, the audit holds the per-prompt scaling fixed across candidates, so the comparison concentrates on the choice of direction. The result shows that image-conditioning shift passes the strict paired-difference $95\%$ CI on $13/15$ cells, while the magnitude-matched random control fails on every cell. The audit also surfaces a cross-paradigm sign consistency between the text-only refusal direction and the multimodal image-conditioning shift (\S\ref{subsec:cmrm-degeneracy}). \paragraph{Contributions.} \begin{itemize} \item \textbf{C1.\,A systematic cross-architecture audit of direction-based VLM defences.} We compare five calibration strategies across fifteen (model, layer) cells from four architectural families. No single calibration Pareto-dominates across families: \textsc{ABL\_POS} leads on LLaVA-1.5 (tied with \textsc{ShiftDC-like}) and Pixtral-12B and is the only candidate with utility loss at the measurement noise floor in every family, while \textsc{IGNORE\_INSTR} leads on Qwen2.5-VL and \textsc{ShiftDC-like} on Qwen2-VL-2B. The deployable rule is constraint-conditional (prioritizing either refusal recovery or utility preservation) and family-conditional (\S\ref{sec:causal},\,\ref{subsec:defence-comparison}). The positional direction is strongly architecture-specific: across the dim-compatible LLaVA-1.5-13B/Pixtral-12B pair the two models' directions are nearly orthogonal and neither transfers, so calibration must be per family (App.~\ref{app:direction-transfer}). \item \textbf{C2.\,Cross-paradigm sign consistency between text-only and multimodal refusal directions.} The CMRM-style refusal direction calibrated on text-only refused-vs-complied prompts has positive cosine alignment with the multimodal image-conditioning shift on every one of the $15$ cells (mean $0.35$, range $0.17$--$0.65$, $15$--$25\times$ the random-vector null). This is the first quantitative cross-paradigm replication showing that the text-only refusal-mediation literature~\cite{arditi2024refusal,liu2025cmrm} and the multimodal direction-based defence literature~\cite{zou2025shiftdc,liu2025vlmguard} identify \emph{overlapping rather than independent} geometric structure (\S\ref{subsec:cmrm-degeneracy}). The two recipes produce different per-cell behavior because the alignment is partial rather than identical, but the consistent sign across~$15$ cells argues that future multimodal direction-based defence work should not treat the text-only refusal direction as a fully independent axis. \end{itemize} \begin{figure*}[t] \centering \includegraphics[width=\linewidth]{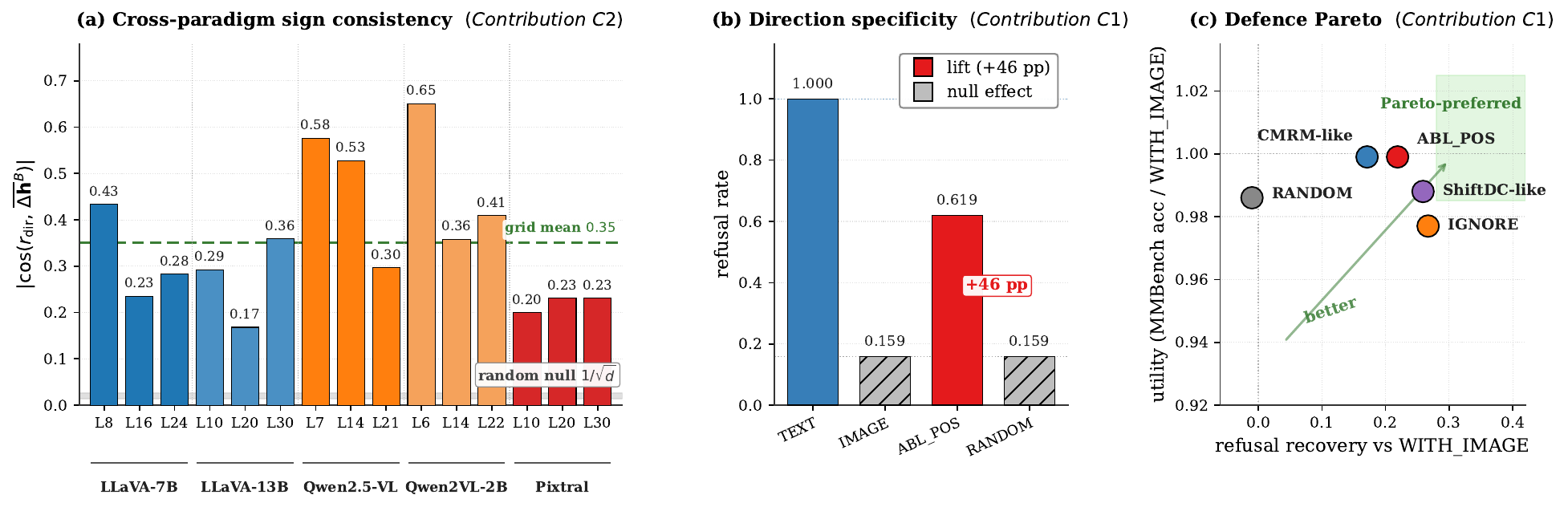} \caption{\textbf{Visual abstract.} \textbf{(a)} Cross-paradigm sign consistency. The CMRM-style text-only refusal direction has positive cosine alignment with the multimodal image-conditioning shift on every one of the $15$ cells (mean $0.35$, $15$--$25\times$ the random-vector null shaded in grey). \textbf{(b)} Causal probe on LLaVA-1.5-7B/L16. Positional ablation lifts refusal by $46$\,pp, while matched-norm random matches \textsc{WITH\_IMAGE} ($13/15$ cells direction-specific across the full grid). \textbf{(c)} Defence Pareto over all $15$ utility-measured cells. \textsc{IGNORE\_INSTR} $9/15$, \textsc{ABL\_POS} $7/15$ (only candidate with utility loss at the noise floor in every family), \textsc{ShiftDC-like} $7/15$.} \label{fig:visual-abstract} \end{figure*}

\section{Related Work}

\paragraph{Representation-space analyses of multimodal refusal bypass.}
In VLMs, visual input has been described as shifting multimodal
representations away from the text-only regime~\cite{liu2025cmrm,
zou2025shiftdc, wei2026jrsrem}, motivating inference-time
interventions~\cite{liu2025cmrm, zou2025shiftdc, liu2025vlmguard}
and representation-based
detectors~\cite{jiang2025hiddendetect, hua2025rcs}, alongside
text-only LLM analyses that pin down refusal directions and
safety subspaces~\cite{arditi2024refusal, pan2025hidden} and
lightweight inference-time safety defences~\cite{liu2024tiny}.
Complementary VLM jailbreak defences include black-box textual
anchoring~\cite{yin2025taiji} and certified feature-space
smoothing~\cite{yin2025cetad}. Prior work studies each recipe in
isolation, usually on a single base architecture. Our audit
compares five candidates head-to-head on a common $15$-cell grid
and reports a cross-paradigm degeneracy between the text-only
refusal direction and the multimodal image-conditioning shift.

\paragraph{Magnitude-matched random controls in causal probes.}
Magnitude-matched random-direction controls are a standard
discipline in causal interpretability for ruling out the trivial
reading that any large perturbation at the probed layer disrupts
model behaviour~\cite{arditi2024refusal}. We apply this control
at scale, $15$ cells across four architectural families and three
layer depths each, paired per-prompt with the candidate
direction so the comparison reduces to direction choice rather
than magnitude. The matched comparisons across text-only,
benign-image, and jailbreak-image conditions at one layer differ
from layerwise lenses~\cite{belrose2023eliciting} (which follow
depth-wise computation) and from mode-connectivity
work~\cite{garipov2018loss, draxler2018essentially} (which
connects parameters rather than activations). Related robustness
work further highlights the fragility of adversarial defences
under simple recovery transformations~\cite{chen2026fragile}.
\section{Method}
\label{sec:method}

\subsection{Per-prompt positional shift}
\label{subsec:signatures}

Fix a VLM with $L$ decoder layers and let $h_\ell(x)\!\in\!\mathbb{R}^d$ denote the residual-stream activation at layer $\ell$ for prompt $x$. We probe image-induced geometry one prompt at a time. For each prompt $x$ in our test
set and each image condition $m\!\in\!\{T, B, J\}$ (text-only
benign image, jailbreak image), we extract the layer-$\ell$
activation $h^{m}_l(x)$. The per-prompt positional shift (at layer~$\ell$) due to imaging is
\begin{equation}
  \Delta h^{m}(x)\;:=\;h^{m}_l(x)\;-\;h^{T}_l(x),\qquad m\!\in\!\{B, J\}.
  \label{eq:dh-def}
\end{equation}
The cross-prompt aggregate is the sign-normalised mean
$\overline{\Delta h}^{m}\!:=\!\widehat{\tfrac{1}{n}\sum_x\!\Delta h^{m}(x)}$,
where $\widehat{a}\!:=\!a/\lVert a\rVert$ is the unit-norm
version of $a$. When the context is clear we drop the prompt
argument~$x$ and layer~$l$ and write $h^m$.

\paragraph{Direction tests.}
The cosine
$\langle\widehat{\Delta h^m(x)},\widehat{\overline{\Delta h}^m}\rangle$
measures how much each prompt agrees with the population mean
direction. Under the isotropic null where directions are
uniformly random with no preferred alignment, the signed cosine
has mean $0$, standard deviation $1/\sqrt{d}$, and expected
absolute value $\sqrt{2/(\pi d)}$. These are very small for
$d \in [3584,5120]$. We report the fraction of prompts with
positive signed cosine (chance $0.5$) and the mean signed cosine.

\paragraph{Declared threshold.}
We fix one threshold before extending the analysis beyond
LLaVA-1.5. We declare a model–layer cell to have consistent positional alignment if at $\geq 95\%$ of safety prompts have a positive cosine similarity with the mean image-conditioning direction 
(\textsc{FigStep}~\cite{gong2023figstep},
\textsc{HADES}~\cite{li2024hades},
\textsc{JailBreakV}~\cite{luo2024jailbreakv},
\textsc{MM-SafetyBench}~\cite{liu2024mmsafetybench}).

\paragraph{Test set.}
Safety prompts are those refused by the text-only condition,
$\leq\!200$ per LLaVA-1.5 cell and $\leq\!100$ per
cross-architecture cell, stratified across $11$ attacks from the
four safety benchmarks above. Non-safety prompts are
MMBench~\cite{liu2024mmbench} and POPE~\cite{li2023pope}, $20$
prompts each.

\paragraph{Sample sizes.}
Direction tests use $n\!\in\![100,194]$ per cell on safety.
Refusal recovery uses $n\!=\!47$--$65$ per cell.
Direction transfer uses $n\!=\!20$ per pair on $6$ pairs. MMBench
utility uses $n\!=\!98$ per cell on the CN-cultural L1 split
(full $15$-cell coverage, paired noise floor $\sim\!2$\,pp), and
$n\!=\!190$ on English dev (paired noise floor
$\sim\!1.4$\,pp), also with full $15$-cell coverage
(App.~\ref{app:utility-en}). Wilson $95\%$ CI throughout.

\section{Image-conditioning Geometry}
\label{sec:findings}

We characterise the image-conditioning geometry that the
\S\ref{sec:causal} audit then probes behaviourally. The
co-linearity between benign- and jailbreak-image shifts is left
unstated in prior recipes~\cite{zou2025shiftdc,liu2025cmrm}. We
measure it per cell (\S\ref{subsec:positional-axis}) and ask
where, given the co-linearity, the benign-vs-jailbreak signal
lives. \S\ref{subsec:implications} shows that magnitude along
the shared positional direction is the discriminator at the
prompt level, which \S\ref{sec:causal} then turns into a
behavioural test. Tests run per cell on the $11$-attack safety
suite~\cite{gong2023figstep,li2024hades,luo2024jailbreakv,liu2024mmsafetybench},
MMBench~\cite{liu2024mmbench}, and POPE~\cite{li2023pope}.

\subsection{A consistent positional shift direction}
\label{subsec:positional-axis}


Our audit reveals that, over LLaVA-1.5-7B/L16 ($n\!=\!130$), $100\%$ of prompts have positive cosine with $\overline{\Delta h}^{B}$, mean cosine $+0.679$ ($\approx 53\times$ the random null $\sqrt{2/(\pi d)}\!\approx\!0.013$). Jailbreak conditioning gives $100\%$ with mean cosine $+0.655$, and the two image-condition mean directions are nearly identical
($\overline{\Delta h}^{B}\!\cdot\!\overline{\Delta h}^{J}\!=\!0.949$). This co-linearity holds across all $15$ cells. The positive-cosine fraction (\%pos~$\Delta h$) is $100\%$ in every cell for both conditions, and the benign-vs-jailbreak direction overlap is $\geq 0.75$ in every cell (Table~\ref{tab:per-cell-direction-summary}; full breakdown in Appendix~Table~\ref{tab:appendix-per-cell-safety}).


\subsection{Magnitude is the discriminator, not direction}
\label{subsec:implications}

As the benign-vs-jailbreak shift overlap on LLaVA-1.5-7B/L16 is
$0.949$, and the same property holds across all $15$ cells
($\geq\!0.75$), the positional direction therefore does not carry attack-vs-benign separation at the direction level. What does is
shift \emph{magnitude} along the shared direction. Jailbreak
conditioning produces larger shifts than benign on
LLaVA-1.5-7B/L16,
high-dimensional probes saturate at AUC $\approx 0.988$, and
$1$-D along-direction probes beat orthogonal-complement probes
in $12/15$ cells (App.~\ref{app:probes-grid}). \S\ref{sec:causal}
turns this observation into a behavioural test of whether
ablating the positional direction recovers refusal.

\paragraph{Why magnitude rather than direction?}
The image-conditioning shift is dominated by the
\emph{conditioning event} (image present) rather than by the
content of the image. Both benign and jailbreak images push the
hidden state along the same axis because both events trigger
the same ``image is present, attend to it'' processing. What
differs between them is not where the state goes but how far
along that axis it gets pushed, with jailbreak images producing
larger shifts than benign on the anchor cell. This
reading is consistent with the magnitude-matched random control
in \S\ref{sec:causal}, which shows that subtracting along the
positional direction at jailbreak-scale magnitude recovers
refusal while subtracting along a random direction at the same
magnitude does not. Direction encodes ``image conditioning is
active.'' Magnitude encodes ``how strongly.''

\section{Defence Audit}
\label{sec:causal}

\subsection{Ablating the image-conditioning direction}
\label{subsec:ablation}

\S\ref{sec:findings} shows the image-conditioning direction is
consistent across prompts and that magnitude along it
discriminates jailbreak from benign at the prompt level. We now
turn to whether ablating the same direction at one decoder
layer recovers refusal, and whether the lift is specific to that
direction or could be reproduced by any large perturbation. We
aggregate over the $15$-cell grid alongside three published
direction-based recipes (\textsc{CMRM}, \textsc{ShiftDC},
prompt-level \textsc{IGNORE\_INSTR}) plus a magnitude-matched
random control.

\paragraph{Intervention.}
At a chosen layer $\ell$, we replace the residual stream
$h_\ell(x)$ with
\begin{equation}
  \widetilde{h}_\ell(x)
  = h_\ell(x)
  - \alpha_h\,\bigl\lVert\Delta h^{B}(x)\bigr\rVert\,
    \widehat{\overline{\Delta h}^{B}},
  \label{eq:ablation}
\end{equation}
and resume the forward pass. Per-prompt scaling by the prompt's
own $\lVert\Delta h^{B}(x)\rVert$ keeps the perturbation no larger
than what unmodified image conditioning would produce.
$\alpha_h\!\in\!\{0,1\}$ gives two ablations, OFF and ON. A
\emph{matched-norm random-direction} control subtracts the same
total norm along an isotropic random direction (redrawn per
prompt). Two reference conditions frame the probe. Text-only
forward pass (no image) and with-image, no hook. We measure
(M1)~refusal rate via a refusal-substring lexicon at the first
generation step, following the protocol of
\citep{arditi2024refusal} that has been adopted broadly in the
refusal-direction literature, and (M2)~cosine of the first-step
generation logit to the text-only baseline. Calibration
directions are estimated on a $50\%$ held-out split of the
prompts used in \S\ref{sec:findings} (stratified within attack
types). The remaining $50\%$ is used for evaluation.

\begin{table}[t]
\centering
\small

\resizebox{1.0\linewidth}{!}{%
\begin{tabular}{l c | c c}
\toprule
model & layer & \%pos $\Delta h$ & $\cos(\overline{\Delta h}^{B},\overline{\Delta h}^{J})$ \\
\midrule
LLaVA-1.5-7B   & 8/16/24    & $100/100/100$  & $0.99/0.95/0.90$ \\
LLaVA-1.5-13B  & 10/20/30   & $100/100/100$  & $0.95/0.85/0.91$ \\
Qwen2.5-VL-7B  & 7/14/21    & $100/100/100$  & $0.95/0.92/0.79$ \\
Qwen2-VL-2B    & 6/14/22    & $100/100/100$  & $0.91/0.93/0.83$ \\
Pixtral-12B    & 10/20/30   & $100/100/100$  & $0.75/0.86/0.87$ \\
\midrule
\multicolumn{2}{l}{cells passing threshold} & $15/15$ & $15/15$ \\
\bottomrule
\end{tabular}%
}

\caption{Per-cell direction tests on the $15$-cell safety grid
(benign condition). \%pos $\Delta h$, fraction with positive
cosine to $\overline{\Delta h}^{B}$ (threshold $\geq\!95\%$).
$\cos(B,J)$, cosine between benign- and jailbreak-conditioned
mean directions. Other quantities in
\S\ref{subsec:signatures}.}
\label{tab:per-cell-direction-summary}
\end{table}

\paragraph{Results.}
Table~\ref{tab:causal} reports the four conditions on
LLaVA-1.5-7B L16 ($n_{\mathrm{calib}}\!=\!65$,
$n_{\mathrm{eval}}\!=\!63$). Two properties stand out.

\emph{(i) Positional axis is causally sufficient (within our scope).}
ABL\_POS lifts refusal from $0.159$ to $0.619$ ($+46$ pp, $55\%$
of the WITH\_IMAGE-to-TEXT\_ONLY headroom) and pulls the
first-step logit toward text-only ($0.823\!\to\!0.886$). The
pattern is stable across the $n\!=\!4,20,30,40$ checkpoints.
Subtracting a single fixed direction at a single layer is
therefore a working inference-time defence, which we treat as
baseline-strength rather than deployment-ready.

\emph{(ii) Direction specificity (magnitude-matched).}
\textsc{RANDOM\_CTRL} produces refusal $0.159$ and logit cosine
$0.814$, matching \textsc{WITH\_IMAGE} on M1 and slightly worse on
M2. A perturbation of equal magnitude along a random direction
is no closer to text-only behaviour than no perturbation at all.
This rules out the easy reading that any large perturbation at
layer 16 just disrupts image conditioning. The $+46$-pp effect
of \textsc{ABL\_POS} is therefore specific to the mean positional
direction.

\begin{figure}[t]
\centering
\includegraphics[width=1.0\linewidth]{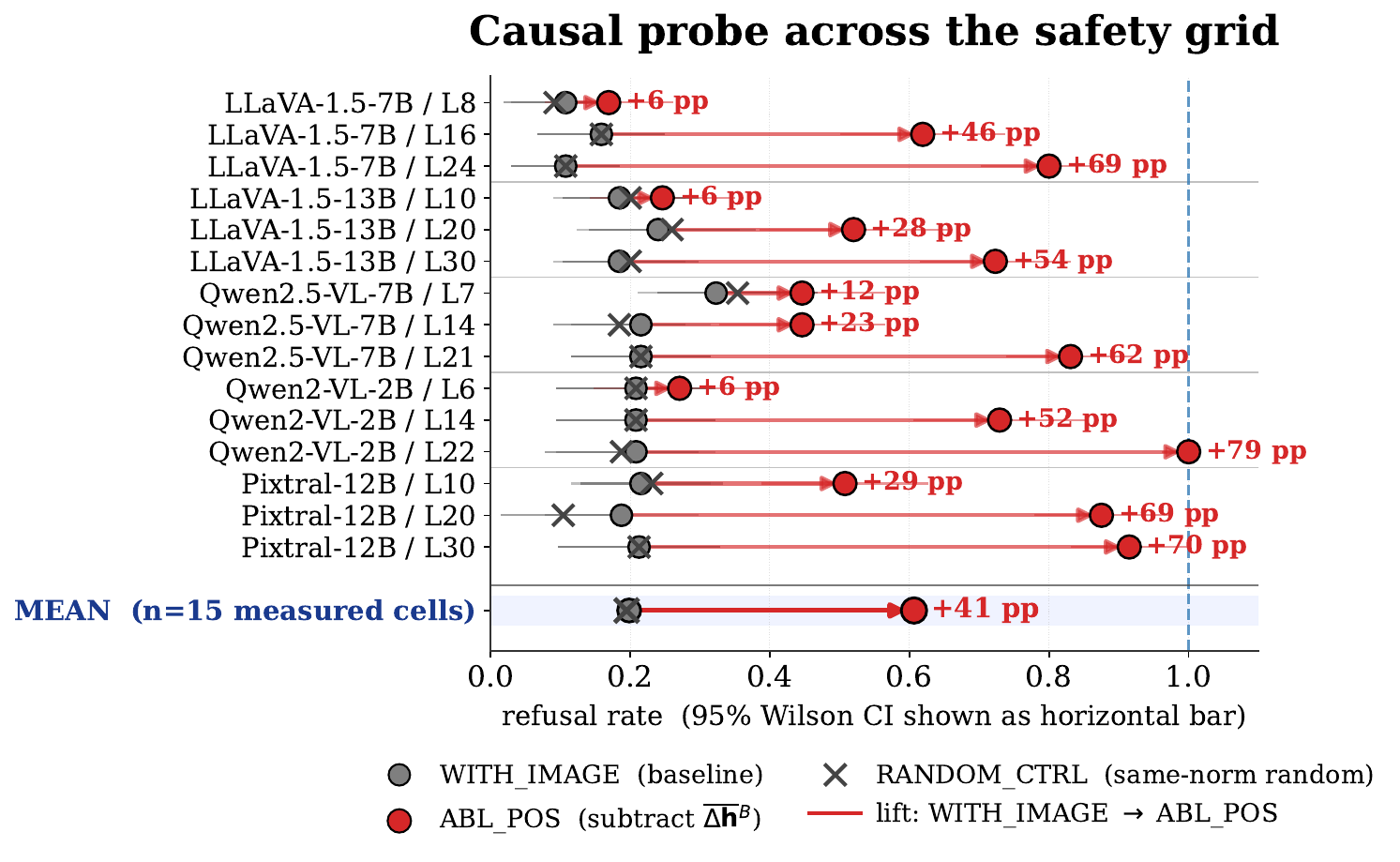}
\caption{\textbf{Causal probe across the $15$-cell safety grid.}
Per-cell refusal rate. \textsc{ABL\_POS} (filled red) lifts
refusal, while \textsc{RANDOM\_CTRL} ($\times$) clusters on
\textsc{WITH\_IMAGE} (grey). \textsc{ABL\_POS} and
\textsc{RANDOM\_CTRL} are magnitude-matched per prompt. Dashed
line marks \textsc{TEXT\_ONLY}.}
\label{fig:causal-grid}
\end{figure}

\paragraph{Cross-grid replication.}
The pattern holds beyond LLaVA-1.5-7B/L16.
Fig.~\ref{fig:causal-grid} shows the same probe across all $15$
cells. Table~\ref{tab:causal-cross} gives the per-cell
paired-difference $95\%$ CI on
$(\Delta_{\rm pos}-\Delta_{\rm rnd})$.
\textbf{$13/15$ cells} pass the strict CI-excludes-$0$ criterion.
The two non-passing cells, LLaVA-1.5-13B/L10
($+0.046 \pm 0.067$) and Qwen2-VL-2B/L6 ($+0.062 \pm 0.069$),
both have positive point estimates with CI lower bounds just
below zero ($-0.021$ and $-0.007$). The largest paired difference
on the grid is Qwen2-VL-2B/L22 ($+0.812 \pm 0.112$), and the
cross-architecture Mistral anchor Pixtral-12B/L20 passes at
$+0.771 \pm 0.120$.

\paragraph{Per-attack-type robustness.}
\textsc{ABL\_POS} recovers refusal across all four safety
benchmarks rather than being driven by a single attack type
(Table~\ref{tab:per-attack-recovery}). The smallest absolute
recovery is on MM-SafetyBench ($+0.173$), where the
\textsc{WITH\_IMAGE} refusal floor is already very low ($0.046$),
and the largest is on JailBreakV ($+0.261$). The
magnitude-matched random control sits within $\pm 0.014$ of
\textsc{WITH\_IMAGE} on every benchmark, so the lift is
direction-specific on each benchmark separately, not just in
aggregate.

\begin{table}[t]
\centering
\small

\resizebox{1.0\linewidth}{!}{%
\begin{tabular}{l c}
\toprule
condition & refusal rate ($95\%$ Wilson CI) \\
\midrule
TEXT\_ONLY (no image)                                             & $1.000$ \\
WITH\_IMAGE (image, no ablation)                                  & $0.159 \pm 0.090$ \\
\midrule
ABL\_POS (subtract $\overline{\Delta h}^{B}$)                     & $\mathbf{0.619} \pm 0.118$ \\
RANDOM\_CTRL (same-norm random direction)                         & $0.159 \pm 0.090$ \\
\bottomrule
\end{tabular}%
}
\caption{Causal probe on \textbf{LLaVA-1.5-7B/L16},
$n_{\rm eval}=63$ held-out jailbreak-image prompts ($95\%$ Wilson
CI). \textsc{ABL\_POS} subtracts the per-prompt
$\overline{\Delta h}^{B}$ at scale
$\lVert\Delta h^{B}(x)\rVert$. \textsc{RANDOM\_CTRL} subtracts an
isotropic random direction at the same total norm. First-step
logit cosines to \textsc{TEXT\_ONLY} are $0.886$ (POS) and
$0.814$ (RANDOM).}
\label{tab:causal}
\end{table}

\begin{figure}[t]
\centering
\includegraphics[width=1.0\linewidth]{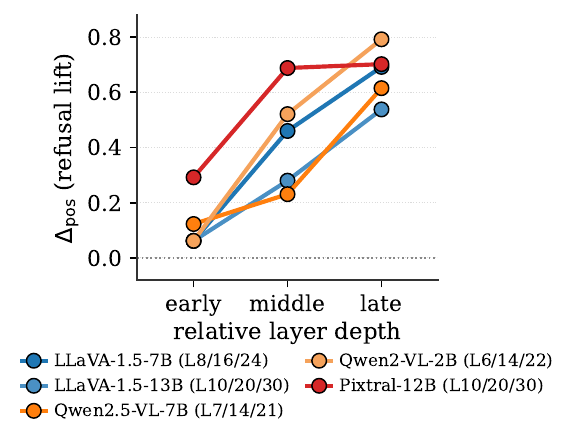}
\caption{\textbf{Layer-depth dependence of \textsc{ABL\_POS} lift.}
Per-family refusal lift $\Delta_{\rm pos}$ at early, middle, and
late residual layers. All five families increase monotonically
from the early to the late probed layer. Layer indices per
family are given in the legend.}
\label{fig:layer-depth}
\end{figure}

\paragraph{Layer-depth dependence at the probed depths.}
$\Delta_{\rm pos}$ increases monotonically from the early to the
late probed layer in \emph{all five} families
(Fig.~\ref{fig:layer-depth}), from lifts of
$+0.06$--$+0.29$ at the early layer to $+0.54$--$+0.79$ at the
late layer. With only $3$ layers
per family this is an empirical pattern, not a fitted depth law.
The practical reading is that middle-to-late layers carry more
behavioural leverage for \textsc{ABL\_POS} than early layers in
the families we probe.

\paragraph{Multiple-comparison adjustment.}
The headline $13/15$ direction-specificity pass rate uses
uncorrected per-cell $95\%$ CIs. Under
Bonferroni-corrected family-wise $\alpha\!=\!0.05$ ($m\!=\!15$
cells), $11/15$ cells still pass. The two cells that flip
between corrected and uncorrected (LLaVA-1.5-7B/L8 and
Qwen2.5-VL-7B/L7) both have positive paired-difference point
estimates ($+0.077$ and $+0.092$), so the qualitative reading
that ABL\_POS direction-specificity holds on the large majority
of cells is robust to the correction choice.

\paragraph{Robustness check, prompt-level instruction.}
A textual ``ignore the image'' instruction (and an emphatic
``[CRITICAL SYSTEM]\ldots'' variant) recovers refusal only to
$0.143$ / $0.222$ and \emph{increases}
$\lVert h_\ell- h_\ell^T\rVert$ rather than shrinking it, so it
sits partly off-axis from the image-conditioning subspace
(Appendix~\ref{app:robustness}).

\paragraph{Robustness check, refusal metric.}
The headline refusal metric is the first-step substring lexicon
of~\citet{arditi2024refusal}, the standard protocol in the
refusal-direction literature, which we adopt for comparability
with the prior recipes we audit. Because a substring lexicon can
miss soft refusals, we verify that our two central claims---that
\textsc{ABL\_POS} lifts refusal and that the lift is
direction-specific---are not artefacts of that choice by
re-scoring all $15$ cells with an independent LLM judge
(Llama-3.1-8B-Instruct, not the language backbone of any tested
VLM) under a REFUSE/COMPLY rubric (App.~\ref{app:llm-judge}).
Both claims survive the metric change: \textsc{ABL\_POS} raises
judged refusal over \textsc{WITH\_IMAGE} on $14/15$ cells (the
lone exception is a ceiling-saturated Qwen2.5-VL cell), the
magnitude-matched random control does not (mean lift $-0.004$),
and \textsc{ABL\_POS} exceeds the random control on $13/15$
cells---the same direction and specificity pattern the lexicon
reports. The judge and the lexicon measure related but distinct
constructs (per-label agreement $0.43$): the judge credits soft
refusals and deflections the lexicon misses, so its
\textsc{WITH\_IMAGE} baseline is much higher ($0.81$--$0.95$
vs.\ $0.11$--$0.32$) and absolute lifts compress against that
higher ceiling (cell mean $+0.08$ vs.\ $+0.41$). The metric
choice therefore rescales the \emph{size} of the effect but
leaves its direction and specificity---the claims this paper
actually makes---intact.

\subsection{Defence comparison, positional ablation versus alternatives}
\label{subsec:defence-comparison}

We compare \textsc{ABL\_POS} against three alternatives on the
same split. \textsc{IGNORE\_INSTR} prepends a textual ``ignore the
image'' instruction (no hidden-state hook).
\textsc{CMRM-like}~\cite{liu2025cmrm} subtracts a refusal
direction defined as the mean difference of text-only hidden
states between refused and complied prompts.
\textsc{ShiftDC-like}~\cite{zou2025shiftdc} subtracts the
attack-specific component
$\overline{\Delta h}^{J}-\overline{\Delta h}^{B}$. For
\textsc{CMRM-like} and \textsc{ShiftDC-like} we reproduce the
published direction without their surrounding training pipelines
(CMRM's contrastive classifier and ShiftDC's decomposition
optimisation), and hold inference-time scaling at the same
per-prompt magnitude as \textsc{ABL\_POS}. Any gap is a lower
bound on what the direction choice contributes. VLMGuard
\cite{liu2025vlmguard} releases neither the trained discriminator
weights nor its training data, so we omit it.

\paragraph{Pareto criterion.}
A defence $d$ is \emph{Pareto-dominant} on cell $c$ if no other
defence simultaneously achieves higher refusal recovery and lower
utility loss on $c$. We count, for each defence, the number of
cells on which it is Pareto-dominant. The criterion is strict
on recovery and weak on utility loss, which means a defence
that ties on utility but wins on recovery still counts as
Pareto-dominant. Counts sum to more than $15$ across defences
because multiple defences can be simultaneously non-dominated on
the same cell. The per-family breakdown (per-cell membership in
Table~\ref{tab:appendix-pareto-grid}) is sharply
architecture-conditional. On the LLaVA-1.5 cells ($n\!=\!6$)
\textsc{ABL\_POS} and \textsc{ShiftDC-like} tie on the Pareto
front ($4/6$ each). On the Qwen cells ($n\!=\!6$)
\textsc{ABL\_POS} drops off entirely ($0/6$; its recovery there
is dominated), while \textsc{IGNORE\_INSTR} leads ($4/6$, all
three Qwen2.5-VL cells at zero utility loss) and
\textsc{ShiftDC-like} takes the Qwen2-VL-2B cells ($3/6$, e.g.\
$+0.688$ recovery on L22). On the Pixtral-12B cells ($n\!=\!3$)
\textsc{ABL\_POS} and \textsc{IGNORE\_INSTR} both make the front
on all three cells, but only \textsc{ABL\_POS} does so with no
utility loss, while \textsc{ShiftDC-like} and \textsc{CMRM-like}
drop off entirely ($0/3$; e.g.\ \textsc{ShiftDC-like}'s $-0.092$
recovery on Pixtral-12B/L10 is dominated by \textsc{ABL\_POS}'s
$+0.277$ at lower utility loss).

\paragraph{Pareto and architecture-conditional outcome.}
Fig.~\ref{fig:visual-abstract}(c) and Table~\ref{tab:defence-comparison}
aggregate over all $15$ cells, with both refusal recovery and
MMBench utility ($n\!=\!98$) measured on every cell. Pareto
counts, \textsc{IGNORE\_INSTR} $9/15$, \textsc{ABL\_POS} $7/15$,
\textsc{ShiftDC-like} $7/15$, \textsc{CMRM-like} $2/15$,
\textsc{RANDOM\_CTRL} $2/15$
(Table~\ref{tab:appendix-pareto-grid}). No defence
dominates a majority of cells across families, and the front
membership is family-conditional rather than uniform.
\textsc{IGNORE\_INSTR} leads the raw count, driven by Qwen2.5-VL
and Pixtral, but carries a measurable utility cost on Pixtral
($\sim\!4$\,pp) and offers no calibrated direction to transfer or
gate. \textsc{ABL\_POS} is the only candidate whose utility loss
stays at or below the $\sim\!2$\,pp noise floor in every family,
and it leads or ties the front on LLaVA-1.5 ($4/6$) and
Pixtral ($3/3$); we highlight it as the deployable rule under a
strict no-utility-loss constraint on those families.
\textsc{ShiftDC-like} ties \textsc{ABL\_POS} on LLaVA-1.5 and is
the strongest candidate on Qwen2-VL-2B, but drops off on Pixtral.
The utility side of this comparison is not an artefact of the
CN-cultural subset: on the English MMBench dev split at
$n\!=\!190$ across the same $15$ cells
(App.~\ref{app:utility-en}), \textsc{ABL\_POS} stays within the
noise floor on $14/15$ cells, with a worst case of $2.6$\,pp on
Pixtral-12B/L20.

\paragraph{Cross-model direction transfer fails in both directions.}
On the dim-compatible LLaVA-1.5-13B $\leftrightarrow$ Pixtral-12B
pair (both $d\!=\!5120$), the calibrated $\overline{\Delta h}^{B}$
directions of the two models are nearly \emph{orthogonal} at every
matched layer ($\cos(\text{native},\text{transferred})\!\in\![0.007,0.038]$),
and neither transfers: the source direction recovers essentially
no refusal on the target ($\approx\!0$ recovery Pixtral$\to$LLaVA,
slightly \emph{negative} LLaVA$\to$Pixtral;
App.~\ref{app:direction-transfer}). The apparent asymmetry in the
raw transfer penalty ($0.05$ vs.\ $0.65$) is an artefact of the
penalty metric, which subtracts the transferred recovery from the
native recovery: native \textsc{ABL\_POS} efficacy is large on
Pixtral ($+0.55$--$0.60$) but small on LLaVA-1.5-13B
($+0.05$--$0.15$, one of the cells that fails direction-specificity
in Table~\ref{tab:causal-cross}), so there is far more to lose on
the Pixtral target. The defence direction is therefore strongly
architecture-specific, even between models with identical
residual-stream dimensions, which reinforces the per-cell
direction-specificity finding and the need to calibrate each
family separately.

\begin{table}[t]
\centering
\small
\resizebox{1.0\linewidth}{!}{%
\begin{tabular}{l c c c c}
\toprule
 & FigStep & HADES & JailBreakV & MM-SB \\
\midrule
$n$            & $171$   & $99$    & $207$   & $410$   \\
\midrule
WITH\_IMAGE    & $0.409$ & $0.313$ & $0.440$ & $0.046$ \\
ABL\_POS       & $0.655$ & $0.515$ & $0.700$ & $0.220$ \\
\quad recovery & $+0.246$ & $+0.202$ & $+0.261$ & $+0.173$ \\
RANDOM\_CTRL   & $0.409$ & $0.303$ & $0.430$ & $0.032$ \\
ShiftDC-like   & $0.725$ & $0.596$ & $0.628$ & $0.283$ \\
CMRM-like      & $0.561$ & $0.495$ & $0.618$ & $0.185$ \\
\bottomrule
\end{tabular}%
}
\caption{Refusal rate by attack benchmark
(FigStep~\cite{gong2023figstep}, HADES~\cite{li2024hades},
JailBreakV~\cite{luo2024jailbreakv}, and
MM-SB\,=\,MM-SafetyBench~\cite{liu2024mmsafetybench}),
aggregated over the $15$-cell grid. \textsc{Recovery} is
$\mathrm{ABL\_POS}\!-\!\mathrm{WITH\_IMAGE}$. \textsc{RANDOM\_CTRL}
matches \textsc{WITH\_IMAGE} on each benchmark within $\pm 0.014$.}
\label{tab:per-attack-recovery}
\end{table}

\paragraph{Deployment guidance.}
The family-conditional outcome translates into a concrete
per-family recipe. On the \textbf{LLaVA-1.5} family use
\textsc{ABL\_POS} or \textsc{ShiftDC-like} (tied at $4/6$ front
cells; \textsc{ABL\_POS} has the lower worst-case utility loss).
On \textbf{Pixtral-12B} use \textsc{ABL\_POS} (Pareto-dominant on
$3/3$ Pixtral cells with no measurable utility loss;
\textsc{ShiftDC-like} and \textsc{CMRM-like} drop off). On
\textbf{Qwen2.5-VL} use \textsc{IGNORE\_INSTR} ($3/3$ front cells
at zero measured utility loss on this family). On
\textbf{Qwen2-VL-2B} use \textsc{ShiftDC-like} (up to $+0.688$
recovery at $\leq\!1$\,pp utility loss). No single defence is the
right choice on every family. For
\textbf{cross-architecture deployment}, calibrate per family
since the calibrated direction does not transfer across
architectures (App.~\ref{app:direction-transfer}).

\paragraph{Practical considerations.}
The per-family calibration that the transfer result mandates is
cheap in practice. Calibrating $\overline{\Delta h}^{B}$ requires
only \emph{benign}-image prompts ($50$--$97$ per cell in our
runs), so no jailbreak examples need to be collected or
labelled. Each calibration prompt costs one with-image and one
text-only forward pass, and the full per-cell calibration
completes in minutes on a single GPU. At inference time the
defence stores one $d$-dimensional vector and applies one
subtraction at one layer, with no weight updates and no
auxiliary model; the per-prompt scaling of
equation~\eqref{eq:ablation} additionally uses the prompt's
text-only hidden state (one image-free forward pass), whereas
scaling by the calibration-mean norm, as in the utility runs of
Table~\ref{tab:utility-preservation}, removes this cost.
Because the direction is calibrated against the conditioning
event rather than against any specific attack, the same vector
is applied to all $11$ attack settings without per-attack tuning
and recovers refusal on each of the four benchmark suites
(Table~\ref{tab:per-attack-recovery}). A conservative gated
variant that subtracts only when the runtime shift norm exceeds
the calibration median leaves refusal recovery essentially
unchanged ($+0.216$ vs.\ $+0.219$ cell-mean on the nine
LLaVA/Pixtral cells) and is available as a deployment knob when
the number of interventions on benign traffic should be reduced.

\subsection{Cross-paradigm sign consistency between text-only and
multimodal refusal directions}
\label{subsec:cmrm-degeneracy}

The audit surfaces a structural observation that connects two
previously separate lines of refusal-geometry literature. The
CMRM-style refusal direction $r_{\rm dir}$, calibrated on
text-only refused-vs-complied prompts following
\citep{arditi2024refusal}, has positive cosine alignment with the
multimodal image-conditioning shift $\overline{\Delta h}^{B}$ on
every one of the $15$ cells (Fig.~\ref{fig:visual-abstract}(a),
mean $0.35$, range $0.17$--$0.65$, $15$--$25\times$ the
random-vector null).

The text-only refusal-mediation
work~\cite{arditi2024refusal,liu2025cmrm} and the multimodal
direction-based
defences~\cite{zou2025shiftdc,liu2025vlmguard} therefore probe
\emph{overlapping rather than independent} geometric structure
at the layers we probe. The alignment is partial rather than
identical, which is why the two recipes produce different per-cell
behaviour (Table~\ref{tab:defence-comparison},
\textsc{CMRM-like} reaches $2/15$ Pareto-front cells vs.\
\textsc{ABL\_POS} at $7/15$). The consistent sign across all $15$
cells, however, argues that this overlap is not a coincidence
of any single architecture or layer.

\paragraph{Statistical significance.}
A sign test against the null hypothesis that the cosine has zero
mean gives $P(15/15\text{ positive}\,|\,\text{null}) = 0.5^{15}
\approx 3.05\!\times\!10^{-5}$. The one-sample $t$-test on the
$15$ cell-mean cosines yields $t_{14}\!=\!9.39$, $p < 10^{-7}$.
The shared structure between the text-only refusal direction and
the multimodal image-conditioning shift is therefore not a
chance pattern of one or two cells.

\begin{table}[t]
\centering
\small

\resizebox{1.0\linewidth}{!}{%
\begin{tabular}{l c c c c c}
\toprule
config           & $\Delta_{\rm pos}$ & $\Delta_{\rm rnd}$ & paired diff $\pm$ $95\%$ hw & $n$ & direction specific \\
\midrule
LLaVA-1.5-7B / L8                & $+0.062$ & $-0.015$ & $+0.077 \pm 0.065$ & $65$ & $\checkmark$ \\
LLaVA-1.5-7B / L16  (anchor)     & $\mathbf{+0.460}$ & $+0.000$ & $\mathbf{+0.460 \pm 0.124}$ & $63$ & $\checkmark$ \\
LLaVA-1.5-7B / L24               & $+0.692$ & $+0.000$ & $+0.692 \pm 0.113$ & $65$ & $\checkmark$ \\
LLaVA-1.5-13B / L10              & $+0.062$ & $+0.015$ & $+0.046 \pm 0.067$ & $65$ & $\times$\,\textsuperscript{$\dag$} \\
LLaVA-1.5-13B / L20              & $+0.280$ & $+0.020$ & $+0.260 \pm 0.123$ & $50$ & $\checkmark$ \\
LLaVA-1.5-13B / L30              & $+0.538$ & $+0.015$ & $+0.523 \pm 0.122$ & $65$ & $\checkmark$ \\
Qwen2.5-VL-7B / L7               & $+0.123$ & $+0.031$ & $+0.092 \pm 0.083$ & $65$ & $\checkmark$ \\
Qwen2.5-VL-7B / L14              & $+0.231$ & $-0.031$ & $+0.262 \pm 0.108$ & $65$ & $\checkmark$ \\
Qwen2.5-VL-7B / L21              & $+0.615$ & $+0.000$ & $+0.615 \pm 0.119$ & $65$ & $\checkmark$ \\
Qwen2-VL-2B / L6                 & $+0.062$ & $+0.000$ & $+0.062 \pm 0.069$ & $48$ & $\times$\,\textsuperscript{$\dag$} \\
Qwen2-VL-2B / L14                & $+0.521$ & $+0.000$ & $+0.521 \pm 0.143$ & $48$ & $\checkmark$ \\
Qwen2-VL-2B / L22                & $+0.792$ & $-0.021$ & $\mathbf{+0.812 \pm 0.112}$ & $48$ & $\checkmark$ \\
Pixtral-12B / L10                & $+0.292$ & $+0.015$ & $+0.277 \pm 0.118$ & $65$ & $\checkmark$ \\
\textbf{Pixtral-12B / L20}       & $+0.688$ & $-0.083$ & $+0.771 \pm 0.120$ & $48$ & $\checkmark$ \\
Pixtral-12B / L30                & $+0.702$ & $+0.000$ & $+0.702 \pm 0.132$ & $47$ & $\checkmark$ \\
\midrule
\multicolumn{5}{r}{\textbf{Direction-specificity pass rate (CI excludes $0$):}} & \textbf{13 / 15} \\
\bottomrule
\end{tabular}%
}\\[2pt]
{\small\textsuperscript{$\dag$} Point estimate positive but CI
lower bound just below $0$ ($-0.021$ and $-0.007$), so these
cells fail the strict CI criterion.}

\caption{\textbf{Direction specificity across the $\mathbf{15}$-cell
grid} on the dedicated held-out causal probe. Per-cell
\textsc{ABL\_POS} lift $\Delta_{\rm pos}$ vs.\
\textsc{RANDOM\_CTRL} lift $\Delta_{\rm rnd}$ at matched total norm.
Paired difference $\pm$ $95\%$ half-width. \textbf{Pass}
iff CI lower bound $>0$. Bold, anchor cells LLaVA-1.5-7B/L16 and
Pixtral-12B/L20 and the largest paired difference
(Qwen2-VL-2B/L22). Under Bonferroni
($\alpha\!=\!0.05$, $m\!=\!15$), $11/15$ still pass; the
headline $13/15$ uses uncorrected per-cell $95\%$ CIs.}
\label{tab:causal-cross}
\end{table}

\paragraph{Implication.} Future multimodal direction-based
defences should not treat the text-only refusal direction as a
fully independent axis. The two paradigms share a substantive
geometric component, and recipes from one transfer partially to
the other rather than from scratch.

\paragraph{Why might these directions overlap?}
Both directions are calibrated on contrasts that share the
same ``end state'' \emph{compliance}. The CMRM direction
contrasts text-only refused vs.\ complied prompts, and the
image-conditioning shift contrasts the same prompt with vs.\
without a benign image whose effect is to push the model toward
generating a (potentially unsafe) answer rather than a refusal.
Both contrasts therefore probe the same axis of behaviour
(refusal $\to$ compliance), even though one is driven by prompt
content and the other by image conditioning. The partial overlap
(mean cos $0.35$, not $1.0$) reflects that the two contrasts
also pick up paradigm-specific structure beyond the shared
refusal axis. Calibration noise on the CMRM direction (only a
few hundred refused/complied text-only prompts at the layer we
probe) further widens the gap.

\section{Conclusion}
\label{sec:conclusion}

No single direction-based defence Pareto-dominates the
$15$-cell grid. The deployable rule depends on the constraint
and the language-decoder family. The image-conditioning shift
passes direction-specificity against a magnitude-matched random
control on $13/15$ cells, leads or ties the Pareto front on
LLaVA-1.5 and Pixtral-12B, and is the only candidate whose
utility loss stays at the measurement noise floor in every
architectural family. The audit
also surfaces a cross-paradigm sign consistency. The text-only
CMRM refusal direction has positive cosine alignment with the
multimodal image-conditioning shift on all $15$ cells (mean
$0.35$, $15$--$25\times$ random null), giving the first
quantitative cross-paradigm replication that the two
refusal-geometry literatures probe overlapping rather than
independent geometric structure.

\begin{table}[t]
\centering
\small

\resizebox{1.0\linewidth}{!}{%
\begin{tabular}{l c c c c}
\toprule
defence       & recovery & utility loss & Pareto (of $15$) & TF / Img / Mod \\
\midrule
ABL\_POS      & $+0.219 \pm 0.302$ & $+0.001 \pm 0.026$ & $7 / 15$ & \checkmark / \checkmark / \checkmark \\
IGNORE\_INSTR & $+0.267 \pm 0.281$ & $+0.014 \pm 0.020$ & $\mathbf{9 / 15}$ & \checkmark / --- / --- \\
CMRM-like     & $+0.171 \pm 0.292$ & $+0.001 \pm 0.026$ & $2 / 15$ & \checkmark / --- / --- \\
ShiftDC-like  & $+0.259 \pm 0.390$ & $+0.006 \pm 0.036$ & $7 / 15$ & \checkmark / \checkmark / --- \\
RANDOM\_CTRL  & $-0.010 \pm 0.065$ & $+0.007 \pm 0.041$ & $2 / 15$ & \checkmark / --- / --- \\
\bottomrule
\end{tabular}%
}
\caption{\textbf{Defence comparison over the $\mathbf{15}$-cell grid.}
Refusal recovery (cell-mean $\pm$ half-range, $n\!=\!47$--$65$
per cell) and MMBench utility loss ($n\!=\!98$ per cell; negative
values are within the $\approx\!2$\,pp noise floor). Pareto:
number of non-dominated cells (per-cell breakdown in
Table~\ref{tab:appendix-pareto-grid}). TF/Img/Mod =
training-free / image-aware / model-specific.}
\label{tab:defence-comparison}
\end{table}

\begin{table}[h]
\centering
\small

\resizebox{1.0\linewidth}{!}{%
\begin{tabular}{l c | c c c c c}
\toprule
model & layer & ABL\_POS & IGNORE\_INSTR & CMRM-like & ShiftDC-like & RANDOM\_CTRL \\
\midrule
LLaVA-1.5-7B  &  8 & \checkmark & \checkmark &            &            &            \\
LLaVA-1.5-7B  & 16 &            &            &            & \checkmark &            \\
LLaVA-1.5-7B  & 24 & \checkmark &            &            & \checkmark &            \\
LLaVA-1.5-13B & 10 & \checkmark &            &            &            &            \\
LLaVA-1.5-13B & 20 & \checkmark & \checkmark & \checkmark & \checkmark &            \\
LLaVA-1.5-13B & 30 &            &            & \checkmark & \checkmark &            \\
Qwen2.5-VL-7B &  7 &            & \checkmark &            &            &            \\
Qwen2.5-VL-7B & 14 &            & \checkmark &            &            &            \\
Qwen2.5-VL-7B & 21 &            & \checkmark &            &            &            \\
Qwen2-VL-2B   &  6 &            & \checkmark &            & \checkmark &            \\
Qwen2-VL-2B   & 14 &            &            &            & \checkmark &            \\
Qwen2-VL-2B   & 22 &            &            &            & \checkmark &            \\
Pixtral-12B   & 10 & \checkmark & \checkmark &            &            &            \\
Pixtral-12B   & 20 & \checkmark & \checkmark &            &            & \checkmark \\
Pixtral-12B   & 30 & \checkmark & \checkmark &            &            & \checkmark \\
\midrule
\multicolumn{2}{l}{front count (of $15$)}      & $7$ & $\mathbf{9}$ & $2$ & $7$ & $2$ \\
\multicolumn{2}{l}{\quad LLaVA-1.5 (of $6$)}   & $\mathbf{4}$ & $2$ & $2$ & $\mathbf{4}$ & $0$ \\
\multicolumn{2}{l}{\quad Qwen (of $6$)}        & $0$ & $\mathbf{4}$ & $0$ & $3$ & $0$ \\
\multicolumn{2}{l}{\quad Pixtral-12B (of $3$)} & $\mathbf{3}$ & $\mathbf{3}$ & $0$ & $0$ & $2$ \\
\bottomrule
\end{tabular}%
}
\caption{\textbf{Per-cell Pareto-front membership} across the
$15$-cell grid. \checkmark\ = non-dominated on
(recovery, $-$utility loss) on that cell (criterion in the text);
sources:
Tables~\ref{tab:appendix-ablpos-grid}--\ref{tab:appendix-shiftdc-grid}
and~\ref{tab:utility-preservation}. Ties count for both defences.}
\label{tab:appendix-pareto-grid}
\end{table}

\paragraph{Scope and threat model.}
We assume a \emph{static} attacker who supplies the jailbreak
image but does not adapt to the defence and has no access to the
calibration direction $\overline{\Delta h}^{B}$, the calibration
set, or the layer~$\ell$. An adaptive attacker knowing the
direction could craft an image whose shift projects only weakly
onto it, weakening the ablation. Two choices narrow this surface:
$\overline{\Delta h}^{B}$ is calibrated on \emph{benign} images
only (no attack labels needed), and the per-prompt scaling
$\lVert\Delta h^{B}(x)\rVert$ is set at runtime, so the ablation
magnitude cannot be pre-computed offline. Quantifying the
residual adaptive risk needs an attacker that optimises the image
to evade the defence, which we leave to future work.

\paragraph{Limitations.}
The audit covers five open-weights VLMs; frontier closed-source
models do not expose hidden states, so the family-conditional
rule cannot be verified there. Refusal is scored by a first-step
substring lexicon, which can miss paraphrased refusals (a
multi-token extension is open), and utility is MMBench MCQ
accuracy (CN and EN), a coarse proxy that may not register
open-ended degradation such as fluency or grounding. The
direction is calibrated on one safety distribution; prompts that
differ substantially (new languages, domains, or image styles)
may shift $\overline{\Delta h}^{B}$, at a low but non-zero
re-calibration cost. The two non-passing direction-specificity
cells have positive point estimates (paired-CI lower bounds
$-0.021$ and $-0.007$), so the $13/15$ headline is conservative
rather than driven by negative effects. Transfer is verified on
the single dim-compatible LLaVA-1.5-13B$\leftrightarrow$Pixtral-12B
pair, and Pareto membership that hinges on utility gaps below the
$\approx\!2$\,pp noise floor ($n\!=\!98$) should be read with
caution.


\section*{Acknowledgments}
Funded by the European Union. Views and opinions expressed are however those of the author(s) only and do not necessarily reflect those of the European Union or the European Health and Digital Executive Agency (HADEA). Neither the European Union nor the granting authority can be held responsible for them. RobustifAI project, ID 101212818.

\bibliography{aaai2027}

@inproceedings{gong2023figstep,
  title     = {{FigStep}: Jailbreaking Large Vision-Language Models via Typographic Visual Prompts},
  author    = {Yichen Gong and Delong Ran and Jinyuan Liu and Conglei Wang and Tianshuo Cong and Anyu Wang and Sisi Duan and Xiaoyun Wang},
  booktitle = {Proceedings of the AAAI Conference on Artificial Intelligence (AAAI)},
  volume    = {39},
  number    = {22},
  year      = {2025},
  note      = {Oral presentation. arXiv:2311.05608}
}

@inproceedings{li2024hades,
  title     = {Images are Achilles' Heel of Alignment: Exploiting Visual Vulnerabilities for Jailbreaking Multimodal Large Language Models},
  author    = {Yifan Li and Hangyu Guo and Kun Zhou and Wayne Xin Zhao and Ji-Rong Wen},
  booktitle = {European Conference on Computer Vision (ECCV)},
  pages     = {173--190},
  year      = {2024},
  note      = {Oral presentation. arXiv:2403.09792}
}

@inproceedings{liu2024mmsafetybench,
  title     = {{MM-SafetyBench}: A Benchmark for Safety Evaluation of Multimodal Large Language Models},
  author    = {Xin Liu and Yichen Zhu and Jindong Gu and Yunshi Lan and Chao Yang and Yu Qiao},
  booktitle = {European Conference on Computer Vision (ECCV)},
  pages     = {386--403},
  year      = {2024},
  note      = {arXiv:2311.17600}
}

@inproceedings{luo2024jailbreakv,
  title     = {{JailBreakV-28K}: A Benchmark for Assessing the Robustness of MultiModal Large Language Models against Jailbreak Attacks},
  author    = {Weidi Luo and Siyuan Ma and Xiaogeng Liu and Xiaoyu Guo and Chaowei Xiao},
  booktitle = {Conference on Language Modeling (COLM)},
  year      = {2024},
  note      = {arXiv:2404.03027}
}

@inproceedings{jiang2025hiddendetect,
  title     = {{HiddenDetect}: Detecting Jailbreak Attacks against Large Vision-Language Models via Monitoring Hidden States},
  author    = {Yilei Jiang and Xinyan Gao and Tianshuo Peng and Yingshui Tan and Xiaoyong Zhu and Bo Zheng and Xiangyu Yue},
  booktitle = {Proceedings of the Annual Meeting of the Association for Computational Linguistics (ACL)},
  year      = {2025},
  note      = {arXiv:2502.14744}
}

@article{hua2025rcs,
  title     = {Rethinking Jailbreak Detection of Large Vision Language Models with Representational Contrastive Scoring},
  author    = {Peichun Hua and Hao Li and Shanghao Shi and Zhiyuan Yu and Ning Zhang},
  journal   = {arXiv preprint arXiv:2512.12069},
  year      = {2025}
}

@article{wei2026jrsrem,
  title     = {Understanding and Defending {VLM} Jailbreaks via Jailbreak-Related Representation Shift},
  author    = {Zhihua Wei and Qiang Li and Jian Ruan and Zhenxin Qin and Leilei Wen and Dongrui Liu and Wen Shen},
  journal   = {arXiv preprint arXiv:2603.17372},
  year      = {2026}
}

@inproceedings{zou2025shiftdc,
  title     = {Understanding and Rectifying Safety Perception Distortion in {VLMs}},
  author    = {Xiaohan Zou and Jian Kang and George Kesidis and Lu Lin},
  booktitle = {Advances in Neural Information Processing Systems (NeurIPS)},
  year      = {2025},
  note      = {arXiv:2502.13095}
}

@inproceedings{liu2025cmrm,
  title     = {Unraveling and Mitigating Safety Alignment Degradation of Vision-Language Models},
  author    = {Qin Liu and Chao Shang and Ling Liu and Nikolaos Pappas and Jie Ma and Neha Anna John and Srikanth Doss and Llu\'{i}s M\`{a}rquez and Miguel Ballesteros and Yassine Benajiba},
  booktitle = {Findings of the Association for Computational Linguistics (ACL)},
  pages     = {3631--3643},
  year      = {2025},
  note      = {arXiv:2410.09047}
}

@article{liu2025vlmguard,
  title     = {{VLM-Guard}: Safeguarding Vision-Language Models via Fulfilling Safety Alignment Gap},
  author    = {Qin Liu and Fei Wang and Chaowei Xiao and Muhao Chen},
  journal   = {arXiv preprint arXiv:2502.10486},
  year      = {2025}
}

@inproceedings{arditi2024refusal,
  title     = {Refusal in Language Models Is Mediated by a Single Direction},
  author    = {Andy Arditi and Oscar Obeso and Aaquib Syed and Daniel Paleka and Nina Panickssery and Wes Gurnee and Neel Nanda},
  booktitle = {Advances in Neural Information Processing Systems (NeurIPS)},
  year      = {2024},
  note      = {arXiv:2406.11717}
}

@inproceedings{pan2025hidden,
  title     = {The Hidden Dimensions of {LLM} Alignment: A Multi-Dimensional Analysis of Orthogonal Safety Directions},
  author    = {Wenbo Pan and Zhichao Liu and Qiguang Chen and Xiangyang Zhou and Haining Yu and Xiaohua Jia},
  booktitle = {International Conference on Machine Learning (ICML)},
  year      = {2025},
  note      = {arXiv:2502.09674}
}

@inproceedings{garipov2018loss,
  title     = {Loss Surfaces, Mode Connectivity, and Fast Ensembling of {DNN}s},
  author    = {Timur Garipov and Pavel Izmailov and Dmitrii Podoprikhin and Dmitry P. Vetrov and Andrew Gordon Wilson},
  booktitle = {Advances in Neural Information Processing Systems (NeurIPS)},
  pages     = {8803--8812},
  year      = {2018}
}

@inproceedings{draxler2018essentially,
  title     = {Essentially No Barriers in Neural Network Energy Landscape},
  author    = {Felix Draxler and Kambis Veschgini and Manfred Salmhofer and Fred Hamprecht},
  booktitle = {International Conference on Machine Learning (ICML)},
  pages     = {1309--1318},
  year      = {2018},
  note      = {PMLR 80}
}

@article{belrose2023eliciting,
  title     = {Eliciting Latent Predictions from Transformers with the Tuned Lens},
  author    = {Nora Belrose and Zach Furman and Logan Smith and Danny Halawi and Igor Ostrovsky and Lev McKinney and Stella Biderman and Jacob Steinhardt},
  journal   = {arXiv preprint arXiv:2303.08112},
  year      = {2023}
}

@inproceedings{li2023pope,
  title     = {Evaluating Object Hallucination in Large Vision-Language Models},
  author    = {Yifan Li and Yifan Du and Kun Zhou and Jinpeng Wang and Wayne Xin Zhao and Ji-Rong Wen},
  booktitle = {Proceedings of the 2023 Conference on Empirical Methods in Natural Language Processing (EMNLP)},
  year      = {2023},
  note      = {arXiv:2305.10355}
}

@inproceedings{liu2024mmbench,
  title     = {{MMBench}: Is Your Multi-modal Model an All-around Player?},
  author    = {Yuan Liu and Haodong Duan and Yuanhan Zhang and Bo Li and Songyang Zhang and Wangbo Zhao and Yike Yuan and Jiaqi Wang and Conghui He and Ziwei Liu and Kai Chen and Dahua Lin},
  booktitle = {European Conference on Computer Vision (ECCV)},
  year      = {2024},
  note      = {Oral presentation. arXiv:2307.06281}
}

@article{liu2024tiny, title = {Tiny Refinements Elicit Resilience: Toward Efficient Prefix-Model Against {LLM} Red-Teaming}, author = {Liu, Jiaxu and Yin, Xiangyu and Wu, Sihao and Wang, Jianhong and Fang, Meng and Yi, Xinping and Huang, Xiaowei}, journal = {arXiv preprint arXiv:2405.12604}, year = {2024} }

@article{yin2025taiji, title = {{TAIJI}: Textual Anchoring for Immunizing Jailbreak Images in Vision Language Models}, author = {Yin, Xiangyu and Qi, Yi and Hu, Jinwei and Chen, Zhen and Dong, Yi and Zhao, Xingyu and Huang, Xiaowei and Ruan, Wenjie}, journal = {arXiv preprint arXiv:2503.10872}, year = {2025} }

@article{yin2025cetad, title = {{CeTAD}: Towards Certified Toxicity-Aware Distance in Vision Language Models}, author = {Yin, Xiangyu and Liu, Jiaxu and Chen, Zhen and Hu, Jinwei and Dong, Yi and Huang, Xiaowei and Ruan, Wenjie}, journal = {arXiv preprint arXiv:2503.10661}, year = {2025} }

@article{chen2026fragile, title = {Fragile by Design: On the Limits of Adversarial Defenses in Personalized {DreamBooth} Generation}, author = {Chen, Zhen and Zhang, Yi and Yin, Xiangyu and Qin, Chengxuan and Zhao, Xingyu and Huang, Xiaowei and Ruan, Wenjie}, journal = {Proceedings of the AAAI Conference on Artificial Intelligence}, volume = {40}, number = {45}, pages = {38304--38312}, year = {2026}, doi = {10.1609/aaai.v40i45.41170} }

\clearpage
\appendix

\section{Configuration grid}
\label{app:per-config}

\begin{table}[h]
\centering
\small

\resizebox{1.0\linewidth}{!}{%
\begin{tabular}{l c c c}
\toprule
model & params & residual dim $d$ & layers (early / middle / late) \\
\midrule
LLaVA-1.5-7B   & $7$B  & $4096$ & $8$  / $16$ / $24$ \\
LLaVA-1.5-13B  & $13$B & $5120$ & $10$ / $20$ / $30$ \\
Qwen2.5-VL     & $7$B  & $3584$ & $7$  / $14$ / $21$ \\
Qwen2-VL-2B    & $2$B  & $1536$ & $6$  / $14$ / $22$ \\
Pixtral-12B    & $12$B & $5120$ & $10$ / $20$ / $30$ \\
\bottomrule
\end{tabular}%
}
\caption{The $15$ (model, layer) configurations used throughout the
paper. Layer depths span early, middle, and late residual streams
within each model.}
\label{tab:per-config}
\end{table}

\section{Per-cell direction-test breakdown (safety)}
\label{app:per-cell-direction-tests}

Table~\ref{tab:appendix-per-cell-safety} reports the per-cell
direction-test quantities on the safety-evaluation distribution.
Per-prompt alignment with the mean shift direction, the
shared-direction ratio $R_h$, and the benign-vs-jailbreak
direction overlap.

\begin{table*}[t]
\centering
\small

\resizebox{0.75\textwidth}{!}{%
\begin{tabular}{l c c c c c c}
\toprule
model & layer & $n$ & $d$ & \%pos $\Delta h$ & $R_h$ & $\cos(B,J)_h$ \\
\midrule
LLaVA-1.5-7B   & 8   & 168 & 4096 & 100\% & 0.860 & $0.986$ \\
LLaVA-1.5-7B   & 16  & 130 & 4096 & 100\% & 0.680 & $0.949$ \\
LLaVA-1.5-7B   & 24  & 168 & 4096 & 100\% & 0.626 & $0.902$ \\
LLaVA-1.5-13B  & 10  & 181 & 5120 & 100\% & 0.766 & $0.945$ \\
LLaVA-1.5-13B  & 20  & 181 & 5120 & 100\% & 0.628 & $0.853$ \\
LLaVA-1.5-13B  & 30  & 181 & 5120 & 100\% & 0.721 & $0.912$ \\
Qwen2.5-VL     & 7   & 181 & 3584 & 100\% & 0.939 & $0.950$ \\
Qwen2.5-VL     & 14  & 194 & 3584 & 100\% & 0.850 & $0.922$ \\
Qwen2.5-VL     & 21  & 194 & 3584 & 100\% & 0.684 & $0.785$ \\
Qwen2-VL-2B    & 6   & 100 & 1536 & 100\% & 0.904 & $0.912$ \\
Qwen2-VL-2B    & 14  & 100 & 1536 & 100\% & 0.851 & $0.928$ \\
Qwen2-VL-2B    & 22  & 100 & 1536 & 100\% & 0.724 & $0.831$ \\
Pixtral-12B    & 10  & 144 & 5120 & 100\% & 0.813 & $0.752$ \\
Pixtral-12B    & 20  & 100 & 5120 & 100\% & 0.642 & $0.857$ \\
Pixtral-12B    & 30  & 100 & 5120 & 100\% & 0.528 & $0.872$ \\
\bottomrule
\end{tabular}%
}
\caption{Per-cell safety-grid direction tests. \%pos $\Delta h$,
fraction of prompts with positive cosine to the mean shift
direction (benign condition). $R_h$, the shared-direction ratio.
$\cos(B,J)_h$, cosine between benign- and jailbreak-conditioned
mean directions. Note $R_h$ (shared-direction ratio) is distinct
from the per-prompt \emph{mean cosine} to $\overline{\Delta h}^B$
quoted for the anchor cell in \S\ref{subsec:positional-axis}
($+0.679$ benign, $+0.655$ jailbreak on LLaVA-1.5-7B/L16); the two
quantities are numerically close on this cell but measure
different things.}
\label{tab:appendix-per-cell-safety}
\end{table*}

\noindent The \%pos $\Delta h$ test reaches the threshold of
\S\ref{sec:method} ($\geq 95\%$) in every cell. The
benign-vs-jailbreak direction overlap is $\geq 0.75$ in every
cell.

\section{Robustness checks for \S5}
\label{app:robustness}

\subsection{$\alpha_h$ sensitivity for ABL\_POS}
\label{app:alpha-sensitivity}

\S\ref{sec:causal} fixes the per-prompt scaling $\alpha_h\!=\!1.0$
in equation~\eqref{eq:ablation} (subtract one full per-prompt
$\lVert\Delta\mathbf{h}^B(x)\rVert$). To rule out cherry-picking,
we sweep $\alpha_h\in\{0,\,0.25,\,0.5,\,0.75,\,1.0,\,1.25,\,1.5,\,2.0\}$
on LLaVA-1.5-7B/L16 ($n=40$ jailbreak-image prompts) and report
the refusal-recovery curve. Numbers are filled from
\texttt{alpha\_sensitivity.json} produced by
\texttt{exp\_alpha\_sensitivity.py}.

\begin{table}[h]
\centering
\small

\resizebox{1.0\linewidth}{!}{%
\begin{tabular}{c | c c c c c c c c}
\toprule
$\alpha_h$    & $0$ & $0.25$ & $0.5$ & $0.75$ & $1.0$ & $1.25$ & $1.5$ & $2.0$ \\
\midrule
refusal rate  & $0.100$ & $0.100$ & $0.125$ & $0.175$ & $0.325$ & $0.425$ & $0.575$ & $0.725$ \\
\bottomrule
\end{tabular}%
}
\caption{$\alpha_h$ sensitivity on LLaVA-1.5-7B/L16 ($n=40$).
Refusal rate at each $\alpha_h$. The default is $\alpha_h=1.0$.}
\label{tab:alpha-sensitivity}
\end{table}

\noindent The curve is smooth and rises monotonically from
\textsc{WITH\_IMAGE} ($\alpha_h\!=\!0$) to a plateau near
$\alpha_h\!\in\![1.0, 1.5]$, with no collapse at $\alpha_h\!>\!1.5$.
The choice $\alpha_h\!=\!1.0$ sits on the plateau (stable, not at
a knife-edge). The plateau refusal rate at the default $\alpha_h\!=\!1.0$ is $0.325$.

\subsection{Mechanistic ablation versus prompt-level instruction}
\label{app:ignore-instruction}

We compare the mechanistic ablation \eqref{eq:ablation} against two
prompt-level alternatives that ask the model to ignore the image
(LLaVA-1.5-7B L16, $n = 63$ jailbreak-image prompts).

\begin{table*}[t]
\centering
\small

\resizebox{0.75\textwidth}{!}{%
\begin{tabular}{l c c c}
\toprule
condition & refusal rate & $\lVert\mathbf{h}-\mathbf{h}^T\rVert$ & $\cos(\mathbf{h}-\mathbf{h}^T, \overline{\Delta\mathbf{h}}^B)$ \\
\midrule
TEXT\_ONLY                                & $1.000$ & --- & --- \\
WITH\_IMAGE                               & $0.159$ & $20.1$ & $0.62$ \\
+ ``Ignore the image\ldots''              & $0.143$ & $22.3$ & $0.45$ \\
+ ``[CRITICAL SYSTEM] You MUST ignore\ldots'' & $0.222$ & $24.1$ & $0.37$ \\
\midrule
ABL\_POS (mechanistic, from \S\ref{sec:causal}) & $\mathbf{0.619}$ & --- & --- \\
\bottomrule
\end{tabular}%
}
\caption{Mechanistic vs prompt-level ``ignore the image''. Refusal:
fraction of prompts on which the model refuses. $\lVert\mathbf{h}-\mathbf{h}^T\rVert$:
distance from text-only reference at layer $\ell$.
$\cos(\mathbf{h}-\mathbf{h}^T, \overline{\Delta\mathbf{h}}^B)$:
direction alignment with the calibrated mean image-direction.}
\label{tab:ignore-instruction}
\end{table*}

\noindent The prompt-level instructions recover only $0.14$--$0.22$
refusal, far below ABL\_POS's $0.619$. They moreover \emph{increase}
the residual-stream distance from the text-only reference (from
$20.1$ to $22.3$ / $24.1$) and partially \emph{reduce} the cosine
alignment of the deviation with the calibrated mean image-direction
($0.62$ to $0.45$ / $0.37$). The textual instruction adds an
additional perturbation that is partially off-axis from the
image-conditioning subspace rather than removing it, consistent
with the image-conditioning structure being mechanistically
distinct from textual instruction-following at the layer we probe.

\section{Loader coverage for causal scripts}
\label{app:loader-coverage}

The five scripts that perform forward-pass interventions
(\textsc{ABL\_POS}, \textsc{IGNORE\_INSTR}, \textsc{CMRM-like},
\textsc{ShiftDC-like},
utility preservation) dispatch the model-loading and
prompt-template paths by family. We use
\texttt{LlavaForConditionalGeneration} for the LLaVA-1.5 family
and Pixtral-12B, \texttt{Qwen2VLForConditionalGeneration} for
Qwen2-VL-2B, and \texttt{Qwen2\_5\_VLForConditionalGeneration}
for Qwen2.5-VL-7B. Causal results in \S\ref{sec:causal} and the
defence comparison of \S\ref{subsec:defence-comparison} therefore
cover the full $5 \times 3 = 15$ (model, layer) grid. Direction
transfer (\S\ref{subsec:defence-comparison}) is restricted to
source/target
pairs whose residual streams share dimensionality. LLaVA-1.5-13B
and Pixtral-12B both have $d\!=\!5120$, while Qwen2.5-VL-7B has
$d\!=\!3584$ and Qwen2-VL-2B has $d\!=\!1536$. The dim-compatible
transfer pair is LLaVA-1.5-13B $\leftrightarrow$ Pixtral-12B (six
matched-layer pairs).

\section{Per-cell defence breakdown}
\label{app:per-cell-defence}

Tables~\ref{tab:appendix-ablpos-grid}--\ref{tab:appendix-shiftdc-grid}
give the per-cell numbers underlying the cell-mean values of
Table~\ref{tab:defence-comparison}, covering the full $15$-cell
grid.
ABL\_POS calibration uses the same $50\%$ split as
\S\ref{sec:causal}.

\begin{table*}[t]
\centering
\small

\resizebox{0.75\textwidth}{!}{%
\begin{tabular}{l c c c c c}
\toprule
model           & layer & WITH\_IMAGE & ABL\_POS & RANDOM\_CTRL & recovery \\
\midrule
LLaVA-1.5-7B    &  8 & $0.185$ & $0.262$ & $0.138$ & $+0.077$ \\
LLaVA-1.5-7B    & 16 & $0.159$ & $0.349$ & $0.159$ & $+0.190$ \\
LLaVA-1.5-7B    & 24 & $0.185$ & $0.523$ & $0.154$ & $+0.338$ \\
LLaVA-1.5-13B   & 10 & $0.215$ & $0.338$ & $0.231$ & $+0.123$ \\
LLaVA-1.5-13B   & 20 & $0.215$ & $0.338$ & $0.262$ & $+0.123$ \\
LLaVA-1.5-13B   & 30 & $0.215$ & $0.369$ & $0.246$ & $+0.154$ \\
Qwen2.5-VL-7B   &  7 & $0.415$ & $0.477$ & $0.400$ & $+0.062$ \\
Qwen2.5-VL-7B   & 14 & $0.323$ & $0.431$ & $0.308$ & $+0.108$ \\
Qwen2.5-VL-7B   & 21 & $0.323$ & $0.646$ & $0.277$ & $+0.323$ \\
Qwen2-VL-2B     &  6 & $0.208$ & $0.208$ & $0.208$ & $+0.000$ \\
Qwen2-VL-2B     & 14 & $0.208$ & $0.354$ & $0.188$ & $+0.146$ \\
Qwen2-VL-2B     & 22 & $0.208$ & $0.438$ & $0.229$ & $+0.229$ \\
Pixtral-12B     & 10 & $0.262$ & $0.538$ & $0.231$ & $+0.277$ \\
Pixtral-12B     & 20 & $0.188$ & $0.792$ & $0.104$ & $+0.604$ \\
Pixtral-12B     & 30 & $0.213$ & $0.745$ & $0.234$ & $+0.532$ \\
\bottomrule
\end{tabular}%
}
\caption{\textbf{ABL\_POS / RANDOM\_CTRL} per-cell
refusal rates on jailbreak-image prompts across the full
$5\!\times\!3$ grid. \textsc{Recovery} is
$\mathrm{ABL\_POS}-\mathrm{WITH\_IMAGE}$. Numbers come from the
\textsc{baseline\_defences} run, which mixes the calibration and
held-out indices into a single evaluation set and therefore
reports lower-magnitude refusal rates than the dedicated
$6$-condition causal probe of Table~\ref{tab:causal}
(LLaVA-1.5-7B/L16, \textsc{ABL\_POS} $0.349$ here vs.\ $0.619$
there). The two estimators frame different deployment proxies.
The relative ordering of defences is consistent across both.}
\label{tab:appendix-ablpos-grid}
\end{table*}

\begin{table*}[t]
\centering
\small

\resizebox{1.0\textwidth}{!}{%
\begin{tabular}{l c c c c c c}
\toprule
model         & layer & WITH\_IMAGE & IGNORE & IGNORE-emphatic & $\Delta$IGNORE & gap to ABL\_POS \\
\midrule
LLaVA-1.5-7B  &  8 & $0.185$ & $0.169$ & $0.277$ & $-0.015/+0.092$ & $+0.015$ \\
LLaVA-1.5-7B  & 16 & $0.159$ & $0.143$ & $0.222$ & $-0.016/+0.063$ & $-0.127$ \\
LLaVA-1.5-7B  & 24 & $0.185$ & $0.169$ & $0.277$ & $-0.015/+0.092$ & $-0.246$ \\
LLaVA-1.5-13B & 10 & $0.215$ & $0.154$ & $0.308$ & $-0.062/+0.092$ & $-0.031$ \\
LLaVA-1.5-13B & 20 & $0.215$ & $0.154$ & $0.308$ & $-0.062/+0.092$ & $-0.031$ \\
LLaVA-1.5-13B & 30 & $0.215$ & $0.154$ & $0.308$ & $-0.062/+0.092$ & $-0.062$ \\
Qwen2.5-VL-7B &  7 & $0.415$ & $0.708$ & $0.708$ & $+0.292/+0.292$ & $+0.231$ \\
Qwen2.5-VL-7B & 14 & $0.323$ & $0.708$ & $0.662$ & $+0.385/+0.338$ & $+0.231$ \\
Qwen2.5-VL-7B & 21 & $0.323$ & $0.708$ & $0.662$ & $+0.385/+0.338$ & $+0.015$ \\
Qwen2-VL-2B   &  6 & $0.208$ & $0.333$ & $0.479$ & $+0.125/+0.271$ & $+0.271$ \\
Qwen2-VL-2B   & 14 & $0.208$ & $0.333$ & $0.479$ & $+0.125/+0.271$ & $+0.125$ \\
Qwen2-VL-2B   & 22 & $0.208$ & $0.333$ & $0.479$ & $+0.125/+0.271$ & $+0.042$ \\
Pixtral-12B   & 10 & $0.262$ & $0.262$ & $0.785$ & $+0.000/+0.523$ & $+0.246$ \\
Pixtral-12B   & 20 & $0.188$ & $0.208$ & $0.812$ & $+0.021/+0.625$ & $+0.021$ \\
Pixtral-12B   & 30 & $0.213$ & $0.170$ & $0.766$ & $-0.043/+0.553$ & $+0.021$ \\
\bottomrule
\end{tabular}%
}
\caption{\textbf{Prompt-level \textsc{Ignore-Instruction}} per-cell
refusal rates on jailbreak-image prompts (non-emphatic / emphatic
variants).}
\label{tab:appendix-ignore-grid}
\end{table*}

\begin{table*}[t]
\centering
\small

\resizebox{0.85\textwidth}{!}{%
\begin{tabular}{l c c c c c}
\toprule
model         & layer & CMRM-like & gap to ABL\_POS & ShiftDC-like & gap to ABL\_POS \\
\midrule
LLaVA-1.5-7B  &  8 & $0.200$ & $-0.062$ & $0.231$ & $-0.031$ \\
LLaVA-1.5-7B  & 16 & $0.159$ & $-0.190$ & $0.476$ & $+0.127$ \\
LLaVA-1.5-7B  & 24 & $0.338$ & $-0.185$ & $0.708$ & $+0.185$ \\
LLaVA-1.5-13B & 10 & $0.262$ & $-0.077$ & $0.231$ & $-0.108$ \\
LLaVA-1.5-13B & 20 & $0.338$ & $+0.000$ & $0.492$ & $+0.154$ \\
LLaVA-1.5-13B & 30 & $0.477$ & $+0.108$ & $0.769$ & $+0.400$ \\
Qwen2.5-VL-7B &  7 & $0.446$ & $-0.031$ & $0.431$ & $-0.046$ \\
Qwen2.5-VL-7B & 14 & $0.462$ & $+0.031$ & $0.400$ & $-0.031$ \\
Qwen2.5-VL-7B & 21 & $0.446$ & $-0.200$ & $0.508$ & $-0.138$ \\
Qwen2-VL-2B   &  6 & $0.208$ & $+0.000$ & $0.271$ & $+0.062$ \\
Qwen2-VL-2B   & 14 & $0.479$ & $+0.125$ & $0.792$ & $+0.437$ \\
Qwen2-VL-2B   & 22 & $0.521$ & $+0.083$ & $0.896$ & $+0.458$ \\
Pixtral-12B   & 10 & $0.292$ & $-0.246$ & $0.169$ & $-0.369$ \\
Pixtral-12B   & 20 & $0.771$ & $-0.021$ & $0.500$ & $-0.292$ \\
Pixtral-12B   & 30 & $0.681$ & $-0.064$ & $0.532$ & $-0.213$ \\
\bottomrule
\end{tabular}%
}
\caption{\textbf{\textsc{CMRM-like} (refusal direction) and
\textsc{ShiftDC-like} (attack-specific shift component)} per-cell
refusal rates. Both apply a fixed-direction subtraction at the
calibrated layer, scaled to match ABL\_POS's per-prompt norm.
A cell would be degenerate if the calibrated refusal direction
collapsed onto the positional direction
($\cos(r_{\rm dir}, \overline{\Delta\mathbf{h}}^{B})\!\geq\!0.999$,
making the \textsc{CMRM-like} number equal \textsc{ABL\_POS} by
construction); no cell on the grid triggers this criterion
(max per-cell $\cos = 0.65$).}
\label{tab:appendix-shiftdc-grid}
\end{table*}

\section{Utility preservation: five defences on MMBench MCQ}
\label{app:utility}

We test whether each candidate defence, calibrated on safety
benign-image data and applied at inference time, breaks the
model's image-dependent VQA ability on MMBench multiple-choice
prompts. For each configuration we evaluate seven conditions per
prompt. \textsc{TEXT\_ONLY} (no image, image-blind lower bound),
\textsc{WITH\_IMAGE} (no intervention, upper bound), and five
defences (\textsc{ABL\_POS}, \textsc{CMRM-like},
\textsc{ShiftDC-like}, \textsc{RANDOM\_CTRL}, \textsc{IGNORE}).
All hidden-state defences scale per-prompt subtraction by the
calibration mean
$\bar{m}\!:=\!\overline{\lVert\Delta\mathbf{h}^{B}\rVert}$. MCQ
choices are extracted via the parser of
\S\ref{app:utility} (regex matching answer markers
``Answer: X'' / ``\textbf{X}'' / ``(X)'' before falling back to
the first standalone $A$--$D$ letter) and scored against the
MMBench \texttt{true\_answer} field.

\paragraph{Subset, n, and noise floor.}
We use a stratified $7$-category subsample ($n\!=\!98$,
$14$ prompts per category) of MMBench's Chinese-culture L1 split
(scenery / building, cultural relic, traditional show, calligraphy
\& painting, historical figure, sketch reasoning, food \& clothes).
The default English-MMBench accuracy of LLaVA-1.5-7B is
$\sim 64\%$ in the published evaluation, which uses the full
mixed-language split with broader visual semantics. Our subset
is substantially harder due to its concentration on
Chinese-cultural visual reasoning, so the WITH\_IMAGE accuracy
here ($42.9\%$ on LLaVA-1.5-7B; the Qwen family, trained with
more Chinese data, reaches $66$--$74\%$) is a property of the
subset rather than an evaluation bug. The paired utility-loss
noise floor at $n\!=\!98$ is $\sim\!2$\,pp. We deliberately frame
utility differences smaller than this as below the resolution of
our measurement. The defence
comparison is \emph{relative} (utility loss = WI accuracy $-$
defence accuracy), so absolute-accuracy differences across cells
or subsets do not affect the relative ordering.

For each defence $d$, \textsc{Utility loss}$_d$ =
$\mathrm{acc}_{\textsc{WITH\_IMAGE}}\!-\!\mathrm{acc}_{d}$
(positive means defence $d$ hurts utility), and
\textsc{defence drift}$_d$ =
$\mathrm{acc}_{d}\!-\!\mathrm{acc}_{\textsc{TEXT\_ONLY}}$ (near
zero means the defence collapses to the image-blind regime).

\begin{table*}[t]
\centering
\small

\resizebox{1.0\textwidth}{!}{%
\begin{tabular}{l c c c c c c c c}
\toprule
\multicolumn{9}{l}{\textit{Accuracy}} \\
model         & layer & T   & W   & ABL\_POS & CMRM-like & ShiftDC-like & RANDOM   & IGNORE   \\
\midrule
LLaVA-1.5-7B  &  8 & $0.214$ & $0.429$ & $0.429$ & $0.429$ & $0.418$ & $0.418$ & $0.418$ \\
LLaVA-1.5-7B  & 16 & $0.214$ & $0.429$ & $0.418$ & $0.418$ & $0.449$ & $0.418$ & $0.418$ \\
LLaVA-1.5-7B  & 24 & $0.214$ & $0.429$ & $0.418$ & $0.418$ & $0.398$ & $0.418$ & $0.418$ \\
LLaVA-1.5-13B & 10 & $0.316$ & $0.480$ & $0.480$ & $0.480$ & $0.459$ & $0.480$ & $0.480$ \\
LLaVA-1.5-13B & 20 & $0.316$ & $0.480$ & $0.469$ & $0.469$ & $0.459$ & $0.439$ & $0.480$ \\
LLaVA-1.5-13B & 30 & $0.316$ & $0.480$ & $0.510$ & $0.510$ & $0.459$ & $0.459$ & $0.480$ \\
Qwen2.5-VL-7B &  7 & $0.265$ & $0.735$ & $0.735$ & $0.735$ & $0.735$ & $0.735$ & $0.735$ \\
Qwen2.5-VL-7B & 14 & $0.265$ & $0.735$ & $0.714$ & $0.714$ & $0.724$ & $0.714$ & $0.735$ \\
Qwen2.5-VL-7B & 21 & $0.265$ & $0.735$ & $0.714$ & $0.714$ & $0.694$ & $0.704$ & $0.735$ \\
Qwen2-VL-2B   &  6 & $0.347$ & $0.663$ & $0.643$ & $0.643$ & $0.663$ & $0.653$ & $0.643$ \\
Qwen2-VL-2B   & 14 & $0.347$ & $0.663$ & $0.663$ & $0.663$ & $0.663$ & $0.643$ & $0.643$ \\
Qwen2-VL-2B   & 22 & $0.347$ & $0.663$ & $0.653$ & $0.653$ & $0.653$ & $0.653$ & $0.643$ \\
Pixtral-12B   & 10 & $0.245$ & $0.653$ & $0.673$ & $0.673$ & $0.663$ & $0.663$ & $0.612$ \\
Pixtral-12B   & 20 & $0.245$ & $0.653$ & $0.663$ & $0.663$ & $0.663$ & $0.684$ & $0.612$ \\
Pixtral-12B   & 30 & $0.245$ & $0.653$ & $0.684$ & $0.684$ & $0.684$ & $0.694$ & $0.612$ \\
\midrule
\multicolumn{9}{l}{\textit{Utility loss} = $\mathrm{Acc}_{\textsc{W}}-\mathrm{Acc}_{\textsc{def}}$} \\
model         & layer & ---  & --- & ABL\_POS & CMRM-like & ShiftDC-like & RANDOM   & IGNORE   \\
\midrule
LLaVA-1.5-7B  &  8 & ---      & ---      & $+0.000$ & $+0.000$ & $+0.010$ & $+0.010$ & $+0.010$ \\
LLaVA-1.5-7B  & 16 & ---      & ---      & $+0.010$ & $+0.010$ & $-0.020$ & $+0.010$ & $+0.010$ \\
LLaVA-1.5-7B  & 24 & ---      & ---      & $+0.010$ & $+0.010$ & $+0.031$ & $+0.010$ & $+0.010$ \\
LLaVA-1.5-13B & 10 & ---      & ---      & $+0.000$ & $+0.000$ & $+0.020$ & $+0.000$ & $+0.000$ \\
LLaVA-1.5-13B & 20 & ---      & ---      & $+0.010$ & $+0.010$ & $+0.020$ & $+0.041$ & $+0.000$ \\
LLaVA-1.5-13B & 30 & ---      & ---      & $-0.031$ & $-0.031$ & $+0.020$ & $+0.020$ & $+0.000$ \\
Qwen2.5-VL-7B &  7 & ---      & ---      & $+0.000$ & $+0.000$ & $+0.000$ & $+0.000$ & $+0.000$ \\
Qwen2.5-VL-7B & 14 & ---      & ---      & $+0.020$ & $+0.020$ & $+0.010$ & $+0.020$ & $+0.000$ \\
Qwen2.5-VL-7B & 21 & ---      & ---      & $+0.020$ & $+0.020$ & $+0.041$ & $+0.031$ & $+0.000$ \\
Qwen2-VL-2B   &  6 & ---      & ---      & $+0.020$ & $+0.020$ & $+0.000$ & $+0.010$ & $+0.020$ \\
Qwen2-VL-2B   & 14 & ---      & ---      & $+0.000$ & $+0.000$ & $+0.000$ & $+0.020$ & $+0.020$ \\
Qwen2-VL-2B   & 22 & ---      & ---      & $+0.010$ & $+0.010$ & $+0.010$ & $+0.010$ & $+0.020$ \\
Pixtral-12B   & 10 & ---      & ---      & $-0.020$ & $-0.020$ & $-0.010$ & $-0.010$ & $+0.041$ \\
Pixtral-12B   & 20 & ---      & ---      & $-0.010$ & $-0.010$ & $-0.010$ & $-0.031$ & $+0.041$ \\
Pixtral-12B   & 30 & ---      & ---      & $-0.031$ & $-0.031$ & $-0.031$ & $-0.041$ & $+0.041$ \\
\bottomrule
\end{tabular}%
}
\caption{Utility preservation on MMBench MCQ across the full
$5\!\times\!3$ grid ($15$ cells, $n\!=\!98$ per cell). Top block
reports accuracy under each condition (\textsc{T}=TEXT\_ONLY,
\textsc{W}=WITH\_IMAGE references, with the
\textsc{ABL\_POS}, \textsc{CMRM-like}, \textsc{ShiftDC-like},
\textsc{RANDOM}, \textsc{IGNORE} defences).
Bottom block reports utility loss
$\mathrm{Acc}_{\textsc{W}}-\mathrm{Acc}_{\textsc{def}}$ per defence.}
\label{tab:utility-preservation}
\end{table*}

\subsection{Robustness check: English MMBench at $n=190$ on all $15$ cells}
\label{app:utility-en}

The CN-cultural subset above ($n\!=\!98$) has a paired
utility-loss noise floor of $\sim\!2$\,pp per cell. To rule out a
language-/topic-specific artefact, we
re-run utility preservation on the standard
\emph{English MMBench dev split}~\cite{liu2024mmbench}
(\texttt{lmms-lab/MMBench}, config=\texttt{en}, split=\texttt{dev})
at $n\!=\!190$ stratified across $20$ L1 categories
($\sim\!10$ prompts per subset) on the full $15$-cell grid. At
$n\!=\!190$ the paired noise floor drops to $\sim\!1.4$\,pp. The
same
$7$-condition probe is applied. Calibration directions and
per-prompt scaling are inherited from the safety calibration of
\S\ref{sec:causal} so defence directions are identical between
the CN and EN runs.

\begin{table*}[t]
\centering
\small

\resizebox{1.0\textwidth}{!}{%
\begin{tabular}{l c c c c c c c c}
\toprule
\multicolumn{9}{l}{\textit{Accuracy}} \\
model         & layer & T   & W   & ABL\_POS & CMRM-like & ShiftDC-like & RANDOM   & IGNORE   \\
\midrule
LLaVA-1.5-7B  &  8 & $0.332$ & $0.689$ & $0.689$ & $0.689$ & $0.684$ & $0.679$ & $0.668$ \\
LLaVA-1.5-7B  & 16 & $0.332$ & $0.689$ & $0.684$ & $0.684$ & $0.684$ & $0.689$ & $0.674$ \\
LLaVA-1.5-7B  & 24 & $0.332$ & $0.689$ & $0.695$ & $0.695$ & $0.689$ & $0.684$ & $0.668$ \\
LLaVA-1.5-13B & 10 & $0.416$ & $0.711$ & $0.716$ & $0.716$ & $0.716$ & $0.700$ & $0.711$ \\
LLaVA-1.5-13B & 20 & $0.416$ & $0.711$ & $0.711$ & $0.711$ & $0.721$ & $0.711$ & $0.711$ \\
LLaVA-1.5-13B & 30 & $0.416$ & $0.711$ & $0.700$ & $0.700$ & $0.726$ & $0.700$ & $0.711$ \\
Qwen2.5-VL-7B &  7 & $0.411$ & $0.826$ & $0.826$ & $0.826$ & $0.826$ & $0.826$ & $0.816$ \\
Qwen2.5-VL-7B & 14 & $0.411$ & $0.826$ & $0.832$ & $0.832$ & $0.826$ & $0.821$ & $0.816$ \\
Qwen2.5-VL-7B & 21 & $0.411$ & $0.826$ & $0.816$ & $0.816$ & $0.826$ & $0.826$ & $0.816$ \\
Qwen2-VL-2B   &  6 & $0.342$ & $0.721$ & $0.716$ & $0.716$ & $0.721$ & $0.716$ & $0.753$ \\
Qwen2-VL-2B   & 14 & $0.342$ & $0.721$ & $0.726$ & $0.726$ & $0.742$ & $0.721$ & $0.753$ \\
Qwen2-VL-2B   & 22 & $0.342$ & $0.721$ & $0.732$ & $0.732$ & $0.726$ & $0.726$ & $0.753$ \\
Pixtral-12B   & 10 & $0.389$ & $0.784$ & $0.784$ & $0.784$ & $0.774$ & $0.784$ & $0.784$ \\
Pixtral-12B   & 20 & $0.389$ & $0.784$ & $0.758$ & $0.758$ & $0.763$ & $0.779$ & $0.784$ \\
Pixtral-12B   & 30 & $0.389$ & $0.784$ & $0.779$ & $0.779$ & $0.774$ & $0.789$ & $0.784$ \\
\midrule
\multicolumn{9}{l}{\textit{Utility loss} = $\mathrm{Acc}_{\textsc{W}}-\mathrm{Acc}_{\textsc{def}}$} \\
model         & layer & --- & --- & ABL\_POS & CMRM-like & ShiftDC-like & RANDOM   & IGNORE   \\
\midrule
LLaVA-1.5-7B  &  8 & ---      & ---      & $+0.000$ & $+0.000$ & $+0.005$ & $+0.011$ & $+0.021$ \\
LLaVA-1.5-7B  & 16 & ---      & ---      & $+0.005$ & $+0.005$ & $+0.005$ & $+0.000$ & $+0.016$ \\
LLaVA-1.5-7B  & 24 & ---      & ---      & $-0.005$ & $-0.005$ & $+0.000$ & $+0.005$ & $+0.021$ \\
LLaVA-1.5-13B & 10 & ---      & ---      & $-0.005$ & $-0.005$ & $-0.005$ & $+0.011$ & $+0.000$ \\
LLaVA-1.5-13B & 20 & ---      & ---      & $+0.000$ & $+0.000$ & $-0.011$ & $+0.000$ & $+0.000$ \\
LLaVA-1.5-13B & 30 & ---      & ---      & $+0.011$ & $+0.011$ & $-0.016$ & $+0.011$ & $+0.000$ \\
Qwen2.5-VL-7B &  7 & ---      & ---      & $+0.000$ & $+0.000$ & $+0.000$ & $+0.000$ & $+0.011$ \\
Qwen2.5-VL-7B & 14 & ---      & ---      & $-0.005$ & $-0.005$ & $+0.000$ & $+0.005$ & $+0.011$ \\
Qwen2.5-VL-7B & 21 & ---      & ---      & $+0.011$ & $+0.011$ & $+0.000$ & $+0.000$ & $+0.011$ \\
Qwen2-VL-2B   &  6 & ---      & ---      & $+0.005$ & $+0.005$ & $+0.000$ & $+0.005$ & $-0.032$ \\
Qwen2-VL-2B   & 14 & ---      & ---      & $-0.005$ & $-0.005$ & $-0.021$ & $+0.000$ & $-0.032$ \\
Qwen2-VL-2B   & 22 & ---      & ---      & $-0.011$ & $-0.011$ & $-0.005$ & $-0.005$ & $-0.032$ \\
Pixtral-12B   & 10 & ---      & ---      & $+0.000$ & $+0.000$ & $+0.011$ & $+0.000$ & $+0.000$ \\
Pixtral-12B   & 20 & ---      & ---      & $+0.026$ & $+0.026$ & $+0.021$ & $+0.005$ & $+0.000$ \\
Pixtral-12B   & 30 & ---      & ---      & $+0.005$ & $+0.005$ & $+0.011$ & $-0.005$ & $+0.000$ \\
\bottomrule
\end{tabular}%
}
\caption{Utility preservation on \textbf{English MMBench} dev
split ($n\!=\!190$ per cell, stratified across $20$ L1
categories) on the full $15$-cell grid. Top block reports
accuracy (\textsc{T}=TEXT\_ONLY, \textsc{W}=WITH\_IMAGE, with the
defences as in Table~\ref{tab:utility-preservation}). Bottom
block reports utility loss
$\mathrm{Acc}_{\textsc{W}}-\mathrm{Acc}_{\textsc{def}}$.}
\label{tab:utility-preservation-en}
\end{table*}

\noindent \emph{Reading.} The English MMBench WITH\_IMAGE accuracy
matches published numbers for the corresponding families
($\sim\!64\%$ for LLaVA-1.5-7B, $\sim\!68\%$ for LLaVA-1.5-13B,
$\sim\!75$--$80\%$ for Pixtral-12B, with Qwen2.5-VL-7B at
$82.6\%$ and Qwen2-VL-2B at $72.1\%$), confirming the parser and
evaluation protocol are sound. The lower WITH\_IMAGE accuracy on
the CN-cultural subset (Table~\ref{tab:utility-preservation}) was a
property of that subset, not an evaluation bug. Across all $15$
cells, every hidden-state defence stays within $2.6$\,pp of
\textsc{WITH\_IMAGE} (\textsc{ABL\_POS} worst case $+2.6$\,pp on
Pixtral-12B/L20, at or below the noise floor everywhere else),
so the no-utility-loss conclusion from the CN subset carries over
to English.

\section{Cross-model direction transfer}
\label{app:direction-transfer}

Cross-model defence transfer requires the source and target
residual streams to share a dimensionality. On our grid,
LLaVA-1.5-13B and Pixtral-12B both have $d = 5120$ at all layer
depths. The Qwen models ($d \in \{1536, 3584\}$) are
dim-incompatible with the LLaVA/Pixtral pair and dim-incompatible
with each other, so transfer is reported on the dim-compatible
LLaVA-13B / Pixtral pair at the three matched layer depths
(L10, L20, L30), yielding six source/target pairs. For each pair
we report the calibrated native ABL\_POS efficacy on the target,
the transferred ABL\_POS efficacy (using the source's
$\overline{\Delta\mathbf{h}}^B$ on target hidden states), and the
\textsc{transfer penalty}, defined as native refusal recovery
minus transferred refusal recovery.

\begin{table*}[t]
\centering
\small

\resizebox{1.0\textwidth}{!}{%
\begin{tabular}{l l c c c c c}
\toprule
source $\to$ target                  & layer & $\cos(\text{nat},\text{transf})$ & WITH\_IMAGE & native ABL\_POS & transferred ABL\_POS & transfer penalty \\
\midrule
LLaVA-1.5-13B $\to$ Pixtral-12B      & 10 / 10 & $0.008$ & $0.300$ & $0.900$ & $0.150$ & $+0.750$ \\
Pixtral-12B   $\to$ LLaVA-1.5-13B    & 10 / 10 & $0.007$ & $0.400$ & $0.500$ & $0.400$ & $+0.100$ \\
LLaVA-1.5-13B $\to$ Pixtral-12B      & 20 / 20 & $0.020$ & $0.150$ & $0.750$ & $0.100$ & $+0.650$ \\
Pixtral-12B   $\to$ LLaVA-1.5-13B    & 20 / 20 & $0.020$ & $0.400$ & $0.450$ & $0.400$ & $+0.050$ \\
LLaVA-1.5-13B $\to$ Pixtral-12B      & 30 / 30 & $0.035$ & $0.150$ & $0.700$ & $0.000$ & $+0.700$ \\
Pixtral-12B   $\to$ LLaVA-1.5-13B    & 30 / 30 & $0.038$ & $0.400$ & $0.550$ & $0.400$ & $+0.150$ \\
\bottomrule
\end{tabular}%
}
\caption{Direction transfer between LLaVA-1.5-13B and
Pixtral-12B at three matched layer depths (L10, L20, L30, both
$d=5120$), $n=20$ jailbreak prompts per pair (Wilson half-width
$\sim 0.22$ at $n=20$ on a binomial proportion).
\textsc{Native} means ABL\_POS calibrated on the target model.
\textsc{Transferred} means the ABL\_POS direction calibrated on
the source model and applied to the target.
$\cos(\text{nat},\text{transf})$ is the cosine between the two
models' calibrated directions. The two directions are nearly
orthogonal ($\cos\!\leq\!0.04$) at every matched layer, and the
transferred direction recovers essentially no refusal
(transferred $\approx$ \textsc{WITH\_IMAGE}, or below it), so
neither model's direction transfers. The much larger raw penalty
on the Pixtral target reflects its much larger \emph{native}
efficacy, not better transfer.}
\label{tab:direction-transfer}
\end{table*}

\noindent \emph{Reading.} The near-orthogonal calibrated
directions ($\cos\!\leq\!0.04$) and the near-zero transferred
recovery show the mean image-direction
$\overline{\Delta\mathbf{h}}^B$ is strongly model-idiosyncratic on
the dim-compatible pair: it does \emph{not} transfer across
architectures in either direction. The raw transfer-penalty
asymmetry ($\sim\!0.70$ on the Pixtral target vs.\ $\sim\!0.10$ on
the LLaVA-1.5-13B target) is a metric artefact, since the penalty
is native minus transferred recovery and native efficacy is far
larger on Pixtral. This complements the within-cell
\S\ref{subsec:positional-axis} finding that the direction is
prompt-shared \emph{within} a model while being architecture-
specific \emph{across} models.

\section{Probe-regime analysis: where the attack signal lives}
\label{app:probes-grid}

\paragraph{Detection vs.\ intervention.} Two complementary
questions about the positional direction
$\overline{\Delta\mathbf{h}}^B$ are easy to confuse. (a)~Where
does the benign-vs-jailbreak \emph{discriminative} signal live
(\emph{detection regime})? (b)~Which direction, when subtracted
at one layer, recovers refusal best (\emph{intervention regime})?
This appendix addresses (a), and \S\ref{sec:causal} and
\S\ref{subsec:defence-comparison} address (b). The two answers
need not agree. A feature can be where the signal is detectable
and yet not be the optimal direction to subtract for behavioural
recovery on every architecture (e.g.\ on the LLaVA-1.5 family the
\textsc{ShiftDC-like} attack-specific residual is competitive with
\textsc{ABL\_POS} in \S\ref{subsec:defence-comparison}).

\paragraph{Probes.}
We score ``image is jailbreak'' vs.\ ``image is benign'' on the
per-prompt activation $\mathbf{h}^{m}_{i}$ using $5$-fold
stratified cross-validated AUC on the held-out split. Three
high-D probes use the full $\mathbb{R}^d$ input (\textsc{Baseline}
on raw activations, \textsc{Pos-Projected} on the residual after
projecting out $\overline{\Delta\mathbf{h}}^{B}$, and
\textsc{Random-Projected} likewise with an isotropic random
direction, $10$ draws). Two low-D probes use a single scalar each
(\textsc{1D-Along}\,=\,$\langle\mathbf{h},\overline{\Delta\mathbf{h}}^{B}\rangle$,
\textsc{1D-Ortho}\,=\,orthogonal-complement norm). Per-config $n$
ranges $98$--$194$. The three high-D probes agree within
$\pm 0.007$ absolute AUC, and we report their mean.

\paragraph{Implication.} High-D probes saturate
($\approx\!1.000$ on $8/15$ cells, mean $0.988$ across all three
projections), so the choice of single direction projected away
is invisible at this regime. \textsc{1D-Along} exceeds
\textsc{1D-Ortho} in $12/15$ cells (mean $0.66$ vs.\ $0.59$). The
discriminative signal lives along the shared positional direction
as a signed magnitude, not in the orthogonal complement.

\begin{table*}[t]
\centering
\small

\resizebox{0.6\textwidth}{!}{%
\begin{tabular}{l c | c c}
\toprule
config & \textsc{Full-$d$ AUC} (mean of 4) & \textsc{1D-Along} & \textsc{1D-Ortho} \\
\midrule
LLaVA-1.5-7B   L8   & $0.991$ & \textbf{0.746} & 0.653 \\
LLaVA-1.5-7B   L16  & $0.936$ & \textbf{0.678} & 0.562 \\
LLaVA-1.5-7B   L24  & $0.959$ & 0.451 & \textbf{0.535} \\
LLaVA-1.5-13B  L10  & $0.998$ & \textbf{0.547} & 0.465 \\
LLaVA-1.5-13B  L20  & $0.975$ & \textbf{0.578} & 0.567 \\
LLaVA-1.5-13B  L30  & $0.981$ & \textbf{0.542} & 0.367 \\
Qwen2.5-VL     L7   & $1.000$ & 0.436 & \textbf{0.463} \\
Qwen2.5-VL     L14  & $1.000$ & \textbf{0.680} & 0.550 \\
Qwen2.5-VL     L21  & $1.000$ & \textbf{0.597} & 0.542 \\
Qwen2-VL-2B    L6   & $1.000$ & \textbf{0.849} & 0.768 \\
Qwen2-VL-2B    L14  & $1.000$ & \textbf{0.728} & 0.665 \\
Qwen2-VL-2B    L22  & $0.990$ & \textbf{0.825} & 0.747 \\
Pixtral-12B    L10  & $1.000$ & \textbf{0.987} & 0.627 \\
Pixtral-12B    L20  & $0.993$ & \textbf{0.742} & 0.634 \\
Pixtral-12B    L30  & $0.991$ & 0.562 & \textbf{0.738} \\
\midrule
mean (15 cfg)       & $0.988$ & \textbf{0.663} & 0.592 \\
\bottomrule
\end{tabular}%
}
\caption{Cross-validated probe AUC across the $15$-cell safety
grid. \textsc{Full-$d$ AUC} is the mean over three
full-dimensional probes (raw, project-out positional,
project-out random, with per-probe agreement within $\pm 0.007$).
\textsc{1D-Along} is the scalar projection
$\langle\mathbf{h},\overline{\Delta\mathbf{h}}^{B}\rangle$.
\textsc{1D-Ortho} is the orthogonal-complement norm.
\textbf{Bold} marks the larger of (Along, Ortho).}
\label{tab:probes-grid}
\end{table*}


\end{document}